%% file: adcf.tex
\documentclass[letterpaper]{article} 
\PassOptionsToPackage{table}{xcolor}
\usepackage[preprint]{aaai2027}  
\usepackage[hyphens]{url}  
\usepackage{graphicx} 
\usepackage{natbib}  
\usepackage{caption} 
\usepackage{algorithm}
\usepackage{algpseudocode}

\usepackage{booktabs}                 
\usepackage{multirow}                 
\usepackage{arydshln}                 

\usepackage{amsmath,amssymb}

\usepackage[utf8]{inputenc}           
\usepackage{microtype}
\usepackage{enumitem}                 

\usepackage{xspace}

\usepackage{xcolor}
\usepackage{hyperref}

\definecolor{linkblue}{RGB}{38, 84, 124}    
\definecolor{citegreen}{RGB}{34, 102, 71}   
\definecolor{urlteal}{RGB}{0, 105, 135}     

\hypersetup{
    colorlinks=true,
    linkcolor=linkblue,
    citecolor=citegreen,
    urlcolor=urlteal
}

\usepackage[noabbrev,capitalize]{cleveref}
\newcommand{\OurMethod}{\textsc{DataMaster}\xspace}
\newcommand{\llm}{G\xspace}

\title{Agentic Instruction Data Selection: Let \emph{\OurMethod} Interpret Your Intent}

\author{
    Fanqi Zhou\textsuperscript{\rm 1}\thanks{Equal contribution.},
    Qiaosheng Chen\textsuperscript{\rm 1}\footnotemark[1],
    Zixian Huang\textsuperscript{\rm 2},
    Gong Cheng\textsuperscript{\rm 1}\thanks{Corresponding author.}
}
\affiliations{
    \textsuperscript{\rm 1}State Key Laboratory of Novel Software Technology, Nanjing University\\
    \textsuperscript{\rm 2}Shanghai Artificial Intelligence Laboratory\\
    221240068@smail.nju.edu.cn, qschen@smail.nju.edu.cn, huangzixian.cn@gmail.com, gcheng@nju.edu.cn
}

\begin{document}
\maketitle

\begin{abstract}
\input{sections/00-abstract}
\end{abstract}

\setcounter{secnumdepth}{1}
\input{sections/01-intro}
\input{sections/02-related-work}
\input{sections/03-method}
\input{sections/04-experiments}

\input{sections/05-conclusion}


\bibliography{custom}

\appendix
\setcounter{secnumdepth}{2}
\input{sections/appendix}

\end{document}

%% file: sections/00-abstract.tex

Although existing instruction data selection methods have introduced various metrics, the inherent complexity of real-world datasets makes it impractical for any single metric to generalize across all scenarios. Developers are thus often forced to manually inspect data and craft heuristic rules for each new application---a tedious and error-prone process. In this paper, we propose a paradigm shift from manual configuration to automated orchestration via the Instruction Data Selection Agent (\OurMethod), which interprets user intent and autonomously composes optimal selection strategies. By allowing users to specify data needs through natural language descriptions, \OurMethod simplifies data curation and removes the burden of manual strategy design. Extensive experiments across the math, medical, and code domains show that \OurMethod outperforms static baselines in most settings and surpasses full-pool training in a substantial number of cases. The implementation of \OurMethod and the scripts needed to reproduce the reported pipeline are publicly available at \url{https://github.com/nju-websoft/DataMaster}.

%% file: sections/01-intro.tex
\section{Introduction}
\label{sec:intro}

The use of instruction data for Supervised Fine-Tuning (SFT) is now a standard stage in the development of Large Language Models (LLMs).
While a vast array of open-source instruction datasets is now available, these collections are often composed of heterogeneous data with diverse distributions that may not align with specific training objectives. 
Indiscriminately utilizing entire datasets not only incurs unnecessary computational overhead but can also lead to suboptimal performance~\citep{zhou2023lima,
chen2024alpagasus,liu2024deita}.

To address these challenges, numerous studies have emerged in recent years focusing on instruction data selection. These works typically concentrate on different dimensions of data value, such as quality~\citep{chen2024alpagasus,chen2025mig}, informativeness~\citep{li2024quantity,li2024superfiltering,liu2024selectit}, and question difficulty together with reasoning-trace properties~\citep{yang2025select2reason}. By designing sophisticated metrics to quantify the value of each sample, these methods aim to select a fixed-size subset consisting of the highest-scoring instances.

\subsubsection{Motivation}
However, these methods typically evaluate data using a \emph{static} selection strategy, whereas the desired subset depends jointly on the heterogeneous composition of the data pool and the user's \emph{dynamic} training needs. In practice, instruction sets often span diverse domains, data characteristics, difficulty levels, and quality tiers, and the same pool must therefore be curated differently as the user's objective changes. Consequently, a monolithic selection strategy cannot reliably isolate the target distribution across these intertwined facets, especially when strengthening a specific model capability.

While some other methods attempt to address the challenge of filtering heterogeneous data by combining multiple metrics~\cite{Rethinking_on_Data_Selection, DEFT}, they still rely on static and hard-coded approaches. We find that even within the same evaluation dimension, different user intentions lead to distinct selection configurations. For example, for the user instructions ``Select  mathematical reasoning data`` and ``Select medical question answering data``, even if both are filtered based on difficulty levels, the former, a reasoning task, may favor samples with moderate difficulty relative to the target model, whereas the latter, a knowledge-based task, may favor harder samples with long-tail knowledge.

\subsubsection{Our Work}
To address this limitation, we propose the Instruction Data Selection Agent (\textbf{\OurMethod}), an agentic data-selection system that translates natural-language user requirements into task-specific selection strategies. Its central novelty lies in \emph{dynamic, intent-conditioned strategy configuration}: rather than applying a fixed metric or a hard-coded combination of metrics, \OurMethod jointly interprets the user's instruction and inspects the candidate data pool to determine how selection should be performed for the current task. We instantiate this principle through four cascaded agents operating over \emph{domain}, \emph{characteristic}, \emph{informativeness}, and \emph{quality}. Each agent configures its selection behavior according to the inferred intent and the data encountered at its stage, progressively isolating the subset most aligned with the requested model capability.

We evaluate \OurMethod on six single-domain training pools spanning math, medical, and code, as well as two multi-domain pools for target-domain selection, using three 7B/8B-scale models as backbones.
In the single-domain experiments, \OurMethod outperforms the static baselines in 16 of the 18 test cases and maintains a consistent lead in the multi-domain experiments.
Using only a natural-language selection instruction, \OurMethod curates a 10K-sample subset for each setting; in the single-domain experiments, training on these subsets outperforms training on the corresponding full, uncurated pool in 12 of the 18 settings. Hosted-LLM call costs average less than US\$$5$ per data-selection task.

%% file: sections/02-related-work.tex

\section{Related Work}
\label{sec:related-work}

\subsubsection{Data Selection for SFT}
\label{sec:rw-selection}

With the rise of LLMs, a growing number of open-source instruction datasets have become available, posing a challenge for developers seeking to efficiently select high-value training data aligned with specific training objectives.
Extensive research has explored data selection schemes across various dimensions, such as domain~\cite{lu2024instag}, statistical characteristics~\cite{cao2023instructmining}, informativeness~\cite{li2024quantity,li2024superfiltering,liu2024selectit}, quality~\cite{chen2025mig}, difficulty~\cite{yang2025select2reason}, and diversity~\cite{ge2024clustering}.
However, given the inherent complexity and diversity of datasets, single-dimensional metrics often struggle to adequately characterize the intricacies of data selection tasks. Consequently, hybrid schemes that integrate multiple dimensions have been widely developed~\cite{du2023mods,zhang2025d3,DEFT}. 
Despite current progress, existing strategies remain largely static, whereas empirical evidence suggests that data selection should be dynamically tailored to task-specific requirements~\cite{Rethinking_on_Data_Selection}.
\emph{Moving beyond static paradigms, \OurMethod combines diverse selection dimensions and uses agents to adapt strategies to the dataset and user intent, enabling automated, task-specific curation.}

\subsubsection{LLM-based Data Selection}
\label{sec:rw-agentic}

Leveraging the rich internal knowledge of LLMs for data selection has emerged as a significant research direction. Specifically, one line of research utilizes the explicit reasoning capabilities of LLMs to perform point-wise quality scoring~\citep{chen2024alpagasus} or pair-wise ranking~\citep{DEFT} of data samples. Another approach employs LLMs to generate semantic domain labels, which are subsequently used for clustering to ensure data diversity~\citep{lu2024instag,teesy}. Additionally, several works exploit implicit signals such as internal gradients to quantify the influence of specific samples on model parameter updates for target downstream tasks~\citep{zhou2023lima,du2023mods}. Despite their reliance on LLMs, these methods remain static and predefined. More recent studies have begun to treat data curation as an agentic process, using agents to orchestrate data discovery, synthesis, and annotation or to optimize broader data-engineering and fine-tuning loops through training and evaluation feedback~\citep{ganguly2025labelingcopilot,DBLP:journals/corr/abs-2603-01712,DBLP:journals/corr/abs-2605-30407,DBLP:journals/corr/abs-2606-04261}. \emph{Unlike these agentic curation systems, which discover or synthesize data and iteratively optimize curation policies through downstream feedback, \OurMethod performs feedback-free selection from an existing SFT pool. It translates a free-form user instruction and direct inspection of the candidate pool into a cascade of domain, characteristic, informativeness, and quality decisions.}





%% file: sections/03-method.tex
\section{Method}
\label{sec:method}

\subsection{Overview}
\label{sec:method-overview}

Given an instruction fine-tuning dataset $\mathcal{D} = \{(x_i, y_i)\}_{i=1}^{n}$ and a target model $\theta$, the goal of instruction data selection is to identify an optimal subset $\mathcal{D}^* \subset \mathcal{D}$ of a target size $k$ ($k \ll n$), such that fine-tuning $\theta$ on $\mathcal{D}^*$ achieves the best performance on specific downstream tasks.

Traditional approaches typically rely on a meticulously designed evaluation function $f(\cdot)$ based on the inherent features of $\mathcal{D}$. However, such customized methods often struggle to generalize across a broad spectrum of tasks with different downstream scenarios.
To adapt to such variability, natural language provides an intuitive and efficient interface for developers. Consequently, we propose incorporating a user-defined instruction $\mathcal{I}$ into the evaluation function to explicitly capture selection intent:
\begin{equation}
    \label{eq:def}
    \mathcal{D}^* = f(\mathcal{I}, \mathcal{D}, \Phi)
\end{equation}
where $\mathcal{I}$ is the natural language instruction expressing the user's intent and $\Phi$ is the shared backbone LLM used to perceive user needs and data characteristics and make selection decisions.
Note that for brevity, the parameters in Equation~(\ref{eq:def}) omit auxiliary tools that may be involved in the selection process, such as a target model $\theta$.
By conditioning the evaluation on $\mathcal{I}$, the function $f$ can dynamically focus the selection strategy on the user's requirements.

Real-world instruction datasets are often large and heterogeneous, making one-pass processing by a single module impractical. We therefore realize the selection function $f$ with four specialized agents that process the data sequentially by \textit{domain}, \textit{characteristic}, \textit{informativeness}, and \textit{quality}:
\begin{equation}
    \mathcal{D}^* = F_{qual} \circ F_{info} \circ F_{char} \circ F_{dom} (\mathcal{I}, \mathcal{D}, \Phi)
\end{equation}
where $F_{dom}, F_{char}, F_{info}$, and $F_{qual}$ represent the filtering or scoring operations performed by the Domain, Characteristic, Informativeness, and Quality agents, respectively. In this framework, each agent dynamically refines the dataset based on the intent $\mathcal{I}$ within its specific functional scope, while the Informativeness Agent additionally probes the target model $\theta$, to eventually produce the target subset $\mathcal{D}^*$.

\begin{figure*}[t]
  \centering
  \includegraphics[page=1,width=\textwidth]{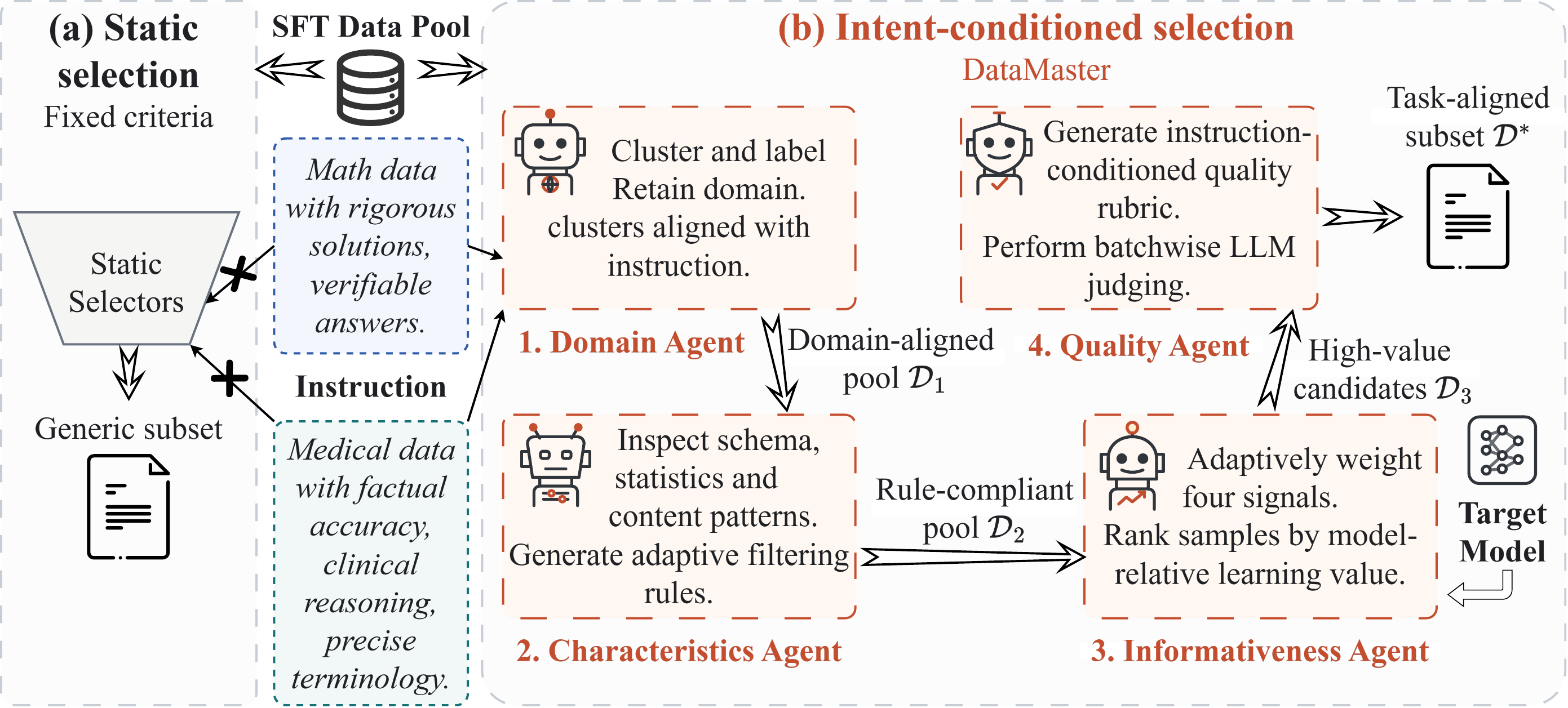}
  \caption{Static selection versus intent-conditioned selection.
  (a) Conventional methods apply fixed selection criteria independently of
  user needs, producing a one-size-fits-all subset. (b) \OurMethod interprets
  the natural-language instruction and the training set to dynamically configure
  a four-agent cascade across domain, characteristics, informativeness, and
  quality. Consequently, the same SFT data pool yields different
  task-aligned subsets for different needs, e.g., mathematical reasoning and medical question
  answering. The target model further informs the learning-value assessment
  of the Informativeness Agent.}
  \label{fig:overview}
\end{figure*}

\Cref{fig:overview} contrasts conventional static selection with our intent-conditioned approach. As shown in \Cref{fig:overview}(a), static selectors ignore task-specific user instructions and apply fixed criteria to the mixed data pool, resulting in a one-size-fits-all subset. In contrast, \Cref{fig:overview}(b) illustrates the overall architecture of \OurMethod, whose four-agent cascade conditions the selection process on the user instruction and produces distinct task-aligned subsets from the same data pool. We next describe the four agents.

\subsection{Stage 1: Domain Agent}

The domain information of the target task is often the most intuitive characteristic for users to describe through natural language. In the first stage of \OurMethod, the Domain Agent interprets the user's requirements within the instruction $\mathcal{I}$. It leaves an already domain-restricted source pool unchanged and otherwise selects matching clusters:
\begin{equation}
    \mathcal{D}_1 = F_{dom}(\mathcal{I}, \mathcal{D}, \Phi) 
    =
    \begin{cases}
        \mathcal{D},
        & \text{already on-domain},\\
        \displaystyle\bigcup_{j \in \mathcal{K}} C_j,
        & \text{otherwise}.
    \end{cases}
\end{equation}
For a multi-domain source pool, the agent embeds the samples and forms semantic clusters, using HDBSCAN when $|\mathcal{D}| \leq n_c$ and KMeans otherwise, where $n_c$ denotes the clustering-switch threshold. Here, $C_j$ denotes the $j$-th resulting cluster, and $\mathcal{K}$ is the set of cluster indices identified by the Domain Agent as matching the instruction $\mathcal{I}$. Full clustering configurations are provided in Appendix~\ref{app:implementation}.

To ensure a precise selection, the agent processes both intra-cluster and inter-cluster information.
For intra-cluster analysis, the agent leverages an LLM to summarize the categorical information for each cluster as follows:
\begin{equation}
    l_j = \llm(P_{lab}, \text{Sample}(C_j), \Phi)
\end{equation}
where $\text{Sample}(\cdot)$ randomly selects $n_s$ representative samples from each cluster to construct the prompt. $P_{lab}$ denotes the prompt template specifically designed to characterize intra-cluster information; the complete template is provided in Appendix~\ref{app:prompts}. The function $\llm(\cdot)$ represents the process of invoking the model $\Phi$ to make inferences, and $l_j$ is a list of one or more category names describing the semantic content of cluster $C_j$, such as "mathematical reasoning" or "medical question answering", for the example in \Cref{fig:overview}(b).

After obtaining the categorical labels $\{l_1, \dots, l_m\}$ for all clusters, the Domain Agent identifies the subset of cluster indices $\mathcal{K}$ that align with the user's intent expressed in $\mathcal{I}$:
\begin{equation}
    \mathcal{K} = \llm(P_{sel}, \mathcal{I}, \{l_1, \dots, l_m \} , \Phi)
\end{equation}
where $P_{sel}$ denotes the cluster-selection prompt template; its complete text is provided in Appendix~\ref{app:prompts}.

\subsection{Stage 2: Characteristic Agent}

Stages 2--4 share a tool-assisted data-inspection mechanism. In addition to their stage-specific selection tools, the agents are equipped with a sandboxed data-analysis interface that allows the backbone LLM to inspect the current data pool through command-line queries and temporary Python scripts. Conditioned on the user instruction $\mathcal{I}$ and the objective of the current stage, the LLM autonomously determines what to inspect, such as previewing a small number of records, querying the dataset size and field schema, or computing lightweight descriptive statistics. These operations do not modify the input data; their observations configure the subsequent selection operation, and no fixed representative subset is constructed in advance. Formally, the observations available to stage $t$ are
\begin{equation}
    \mathcal{O}_t = \operatorname{Inspect}(\mathcal{I}, \mathcal{D}_{t-1}; \Phi, \mathcal{T}_{\mathrm{ana}}), \qquad t \in \{2,3,4\},
\end{equation}
where $\mathcal{T}_{\mathrm{ana}}$ is the shared sandboxed data-analysis tool and $\mathcal{O}_t$ contains observations from model-directed exploration.

Customizing filtering rules based on data characteristics is efficient but often fails to generalize across diverse tasks. To overcome this limitation, the second stage of \OurMethod employs a Characteristic Agent. This agent is designed to automatically analyze data characteristics and dynamically generate appropriate filtering rules, ensuring the selection strategy remains adaptive to varying data distributions.

Given the subset $\mathcal{D}_1$ filtered in Stage~1, the Characteristic Agent selects data by formulating rules that are specifically tailored to both the data characteristics and the user's intent:
\begin{equation} \mathcal{D}_2 = F_{char}(\mathcal{I}, \mathcal{D}_1, \Phi) = \text{Match}(\mathcal{D}_1, \mathcal{R}(\mathcal{I}, \mathcal{D}_1, \Phi)) 
\end{equation}
where $\text{Match}(\cdot, \cdot)$ denotes a rule-matching filter, and $\mathcal{R}(\cdot, \cdot, \cdot)$ represents the rule generator. Based on the user instruction $\mathcal{I}$ and the tool-assisted observations $\mathcal{O}_2$ collected from $\mathcal{D}_1$, the model $\Phi$ synthesizes a set of specific filtering rules:
\begin{equation} \mathcal{R}(\mathcal{I}, \mathcal{D}_1, \Phi) = \llm(P_{char}, \mathcal{I}, \mathcal{O}_2, \Phi) \end{equation}
where $P_{char}$ is the prompt template for rule synthesis; its complete text is provided in Appendix~\ref{app:prompts}.

By integrating user intent with the observed data characteristics, the rule generator adaptively formulates diverse filtering criteria.
These decisions include, but are not limited to, generating specific prohibited keywords, establishing sequence length constraints, and defining linguistic requirements.

\subsection{Stage 3: Informativeness Agent}

After the initial coarse-grained passes, \OurMethod proceeds to a fine-grained selection stage. Here, an Informativeness Agent is utilized to assess the learning value of the remaining samples relative to the target model, ensuring that only the most beneficial data is retained for training:
\begin{equation}
    \mathcal{D}_3 = F_{info}(\mathcal{I}, \mathcal{D}_2, \Phi) = \text{Top-K}(\mathcal{D}_2, S(\mathcal{I}, \mathcal{D}_2, \Phi))
\end{equation}
where $S(\cdot, \cdot, \cdot)$ denotes the scoring function that quantifies the informativeness of each individual sample. The $\text{Top-K}(\cdot, \cdot)$ operator then identifies a subset of $k^*$ samples with the highest scores~($k^* > k$).

Given that traditional metrics such as negative log-likelihood (NLL) and token-level entropy often exhibit inconsistent performance across diverse tasks, the Informativeness Agent introduces an adaptive evaluation strategy. By interpreting the user's intent within $\mathcal{I}$ together with the tool-assisted observations $\mathcal{O}_3$ collected from $\mathcal{D}_2$, the agent dynamically generates a task-specific weight vector $\mathbf{w}=[w^1,\dots,w^{m_1}]$, where $w^j$ controls the contribution of the $j$-th informativeness metric. We then compute the final informativeness score as
\begin{align}
    \begin{split}
        S(\mathcal{I}, \mathcal{D}_2, \Phi) &= \{ \sum_{j=1}^{m_1} w^j \cdot c_i^j \mid d_i \in \mathcal{D}_2 \}\, \\
    [w^1, \dots, w^{m_1}] &= \llm(P_{inf}, \mathcal{I}, \mathcal{O}_3, \Phi) \,
    \end{split}
\end{align}
where $c_i^j$ denotes the value of the $j$-th informativeness metric for sample $i$. The weights are non-negative and normalized such that $\sum_{j=1}^{m_1} w^j=1$. We employ four metrics ($m_1=4$): NLL, token-level entropy, semantic drift, and MeanDiff, adapted from the mean parameter-difference score of ResoFilter~\citep{tu2025resofilter}, which measures the parameter-change magnitude induced by a single-sample update. Definitions of all four metrics appear in Appendix~\ref{app:info_metric}. The $P_{inf}$ prompt appears in Appendix~\ref{app:prompts}.

Through the Informativeness Agent, different metrics are weighted according to task characteristics. For example, NLL is prioritized for knowledge-intensive tasks to detect informational gaps but receives lower weights in reasoning tasks; representative allocations appear in Appendix~\ref{app:inf-weights}.




\subsection{Stage 4: Quality Agent}

Finally, \OurMethod employs a Quality Agent to select the $k$ samples with the highest quality scores to constitute the final training set $\mathcal{D}^*$:
\begin{equation}
    \mathcal{D}^* = F_{qual}(\mathcal{I}, \mathcal{D}_3, \Phi) = \text{Top-K}(\mathcal{D}_3, Q(\mathcal{I}, \mathcal{D}_3, \Phi))
\end{equation}
where $Q(\cdot, \cdot, \cdot)$ is a scoring function implemented based on the LLM-as-a-Judge to evaluate the quality of each sample.

Because quality criteria vary by task, the Quality Agent dynamically formulates a scoring rubric and weights its indicators accordingly. For example, it prioritizes factual correctness for knowledge-intensive tasks and logical rigor for reasoning tasks:
\begin{equation}
    Q(\mathcal{I}, \mathcal{D}_3, \Phi) = \{ \sum_{j=1}^{m_2} z^j \cdot q_i^j \mid d_i \in \mathcal{D}_3 \} \, 
\end{equation}
where $z^j$ represents the weight assigned to the $j$-th metric, and $q_i^j$ denotes the score of the $i$-th sample on that metric.

The scoring rubrics and their associated weights are synthesized through an in-depth interpretation of the user instruction $\mathcal{I}$ and the tool-assisted observations $\mathcal{O}_4$ collected from $\mathcal{D}_3$:
\begin{equation}
    \mathcal{R}_{qual}
    =
    \{(r^j, z^j)\}_{j=1}^{m_2}
    =
    \llm(P_{rub}, \mathcal{I}, \mathcal{O}_4, \Phi)
\end{equation}
where $P_{rub}$ is the fixed rubric-generation prompt template, $r^j$ contains the name and description of the $j$-th judging dimension, and $z^j$ is its weight. The resulting $\mathcal{R}_{qual}$ is therefore generated dynamically for the current instruction and data pool. The complete template is provided in Appendix~\ref{app:prompts}. An example generated rubric is provided in Appendix~\ref{app:implementation}.

Finally, the Quality Agent instantiates the fixed batch-scoring scaffold $P_{cri}$ with $\mathcal{R}_{qual}$ and conducts listwise scoring on each candidate batch $\mathcal{B} \subset \mathcal{D}_3$:
\begin{equation}
    \{[q_i^1, \dots, q_i^{m_2}] \mid d_i \in \mathcal{B} \}
    =
    \llm(P_{cri}(\mathcal{R}_{qual}), \mathcal{B}, \Phi) \,.
\end{equation}
Given that the scale of $\mathcal{D}_3$ may still be substantial, we partition the dataset into multiple batches of size $b$ for parallel evaluation in practice. In our experiments, the same shared backbone LLM $\Phi$ generates the rubric and performs the batched judging; a compatible lightweight judge can instead be used when deployment cost is the priority.

%% file: sections/04-experiments.tex
\section{Experiments}
\label{sec:experiments}

\subsection{Setup}
\label{sec:setup}

\paragraph{Models.}

For the target models in our experiments, we employ three backbones: two base models, Qwen2.5-7B~\cite{qwen2025qwen25} and Llama-3.1-8B~\cite{grattafiori2024llama3}, and one instruction model, Olmo-3-7B-Instruct~\cite{olmo2025olmo3}.
We employ DeepSeek-V4-Flash~\cite{deepseekv4} as the backbone LLM for \OurMethod.


\input{Tables/main_results}

\paragraph{Benchmarks.}
For the single-domain experiments, we evaluate our method on 11 benchmarks across the math, code, and medical domains.
For the \textbf{math} domain, we use six benchmarks: GSM8K, MATH, Minerva-Math, AMC23, and AIME 2024 and 2025.
For the \textbf{code} domain, we use LiveCodeBench, HumanEval, and BigCodeBench.
For the \textbf{medical} domain, we use MedQA and MMLU-Medical.
The multi-domain experiments reuse the math and code benchmarks above and additionally evaluate the science target domain using MMLU-STEM and GPQA-Diamond.
Together, these benchmarks span open-ended reasoning at varied difficulty, executable code generation, and domain-specific multiple-choice reasoning, testing whether selection gains transfer across task formats and evaluators. Benchmark sources are provided in Appendix~\ref{app:benchmark-details}.

\paragraph{Metrics.}
We report all benchmark scores as percentages, with higher values
indicating better performance. For the six math benchmarks, we
extract and normalize the final answer from each generation and report
answer accuracy, accepting either a normalized exact match or parsed
mathematical equivalence to the reference answer. For MedQA,
MMLU-Medical, MMLU-STEM, and GPQA-Diamond, we report multiple-choice
accuracy based on the extracted answer option. For LiveCodeBench,
HumanEval, and BigCodeBench, we generate one completion per problem and
report pass@1, where a completion is counted as correct only if it passes
the benchmark-provided tests. The setting-level scores in
\Cref{tab:main-results,tab:multi-disc} are unweighted arithmetic means of
the percentage scores across the benchmarks in the corresponding target
domain: six for math, two for medical, three for code, and two
for science.

\paragraph{Data Pools.}
For single-domain experiments, we select two public SFT pools per domain that differ in scale and construction to test robustness across sources for a fixed target domain. For the math domain, we use OpenR1-Math (94K) and Synthetic-Math (100K). For the medical domain, we use Medical-Reasoning (20K) and UltraMedical (410K). For the code domain, we use OSS-Instruct (50K) and Code-Alpaca (20K).

For multi-domain experiments, we use OpenHermes-2.5 (1M) and Tulu-3-SFT-Mixture (939K) as two large public mixed-domain SFT pools with overlapping math and code content but different compositions. This design evaluates target-domain selection across distinct source mixtures under the same 10K budget. Given that OpenHermes-2.5 also contains data from the science domain, we additionally evaluate science selection on this dataset.

Representative natural-language instructions for both settings are provided in Appendix~\ref{app:task-instructions}. Dataset sources and licenses appear in Appendix~\ref{app:data-licenses}.

\input{Tables/multi_domain}

\paragraph{Baselines.}

For the single-domain experiments, we include a no-selection control that fine-tunes each target model on the uncurated source \textbf{Full Pool}.

\textbf{Single-domain baselines.} We first include \textbf{Random}, which uniformly samples 10K examples from the source pool without using a selection criterion. We then compare five static selection methods. Specifically, \textbf{SuperFiltering}~\cite{li2024superfiltering} and \textbf{SelectIT}~\cite{liu2024selectit} measure informativeness, \textbf{MIG}~\cite{chen2025mig} measures quality through information gain, and \textbf{Select2Reason}~\cite{yang2025select2reason} jointly ranks question difficulty and reasoning-trace length. We also employ \textbf{DEFT}~\cite{DEFT}, which scores complexity, quality, and knowledge under a diversity constraint.
For the dynamic agent baseline, we use \textbf{Claude Code} with Claude Opus 4.7 as its backend model.

\textbf{Multi-domain baselines.} We compare four static selection methods and a dynamic agent
baseline. For the two-stage static methods, we employ
\textbf{SuperFiltering (Tag)} and \textbf{SuperFiltering (Sim)}, which
select target-domain candidates using source tags and BGE-M3 embedding
similarity to target examples~\citep{bgem3}, respectively, before
SuperFiltering selects the final 10K subset. We also include
\textbf{DSIR}~\citep{xie2023dsir} and
\textbf{TSDS}~\citep{liu2024tsds} as target-guided static baselines
that align the selected data with a reference distribution estimated
from target examples. For the dynamic agent baseline, we use the same
\textbf{Claude Code}, which receives the same multi-domain instruction
$\mathcal{I}$ as \OurMethod, identifies the target domain, and selects the final
subset in one pass.
However, \textbf{Full Pool} is not included in the multi-domain
comparison because training on the entire mixed-domain pool would
depart from the shared 10K selection budget and conflate selection
quality with substantially greater training data and compute.

Implementation notes for all baselines are provided in Appendix~\ref{app:baseline-impl}.

\paragraph{Implementation Details.}
All data selection methods operate under a 10K subset budget.
To ensure comparability, all methods within a given setting follow the
same training and evaluation protocol. 
Implementation details for \OurMethod are provided in Appendix~\ref{app:implementation}.
Training and reproducibility details are provided in
Appendix~\ref{app:reproducibility}.

\subsection{Results on Single-Domain Experiments}
\label{sec:main-results}

\Cref{tab:main-results} compares \OurMethod with seven baselines, comprising six static methods and one dynamic method. \OurMethod ranks first in 13 of 18 settings and remains in the top three in the other five. Per-benchmark single-domain results are provided in Appendix~\ref{app:per-benchmark-single}.

\paragraph{Against Random.}
\OurMethod outperforms Random in all 18 settings, with gains ranging from 3.47 to 9.94 points.

\paragraph{Against the static baselines.}
Across all three target models and domains, \OurMethod achieves the top average score among the six static baselines in 16 of 18 experimental settings.
\OurMethod maintains at least a one-point lead over the strongest single-dimension baseline SuperFiltering in 17 settings, and achieves a similar margin over the multi-dimension baseline DEFT in 13 settings.
Even when compared to the strongest baseline DEFT, \OurMethod still maintains a clear advantage, achieving a 4.40-point improvement in code-related tasks using Olmo-3-7B-Instruct and the OSS-Instruct source pool.
These results support dynamic selection tailored to instructions and data rather than a fixed strategy across scenarios.

\paragraph{Against the dynamic baseline.}
Claude Code, powered by Claude Opus 4.7, generally outperforms the static methods, supporting the value of dynamic agent-based selection. \OurMethod nevertheless achieves a higher average score in 14 of 18 settings, with a 3.07-point lead for Qwen2.5-7B on OpenR1-Math.


\paragraph{Against the Full Pool training.}

To examine whether data selection compromises model performance, we also compare \OurMethod with the Full Pool baseline in \Cref{tab:main-results}, where models are trained on the entire dataset. We observe that although Full Pool training uses $2\times$--$41\times$ as many examples as \OurMethod, it underperforms our approach in 12 of 18 experimental settings. Full Pool achieves higher scores in the remaining six settings. Consequently, our method lowers training costs and often achieves superior performance with substantially fewer training examples.

\paragraph{Practical cost.}
Across the single-domain settings, \OurMethod incurs approximately US\$4 in hosted-LLM API costs per data selection task.


\subsection{Results on Multi-Domain Experiments}
\label{sec:multi-disc-quant}
\paragraph{Results.}
To reflect real-world practices where data is typically selected from mixed-domain pools, we conduct experiments on multi-domain data selection in \Cref{tab:multi-disc}. \OurMethod outperforms all four static baselines in every test case, with gains ranging from 0.33 to 2.84 points over the strongest one. It also surpasses Claude Code, the most competitive baseline, by 0.04 to 1.34 points across all trials. Such consistent outperformance underscores the robustness and effectiveness of \OurMethod in complex selection environments. Per-benchmark multi-domain results are provided in Appendix~\ref{app:per-benchmark-multi}.



\paragraph{Practical cost.}
Across the five multi-domain settings, \OurMethod requires only about US\$5 in API token costs per data selection task.


\subsection{Ablation Study}
\label{sec:ablation}

As Stage~1 of \OurMethod is tailored for multi-domain data selection, we structure the ablation study into two phases. Initially, we investigate the influence of Stage~1 by conducting experiments on multi-domain data pools. Following this, we ablate Stages~2 to~4 on single-domain data pools.

\paragraph{Ablation of Stage~1.}

\begin{figure}[t]
  \centering
  \includegraphics[width=\columnwidth]{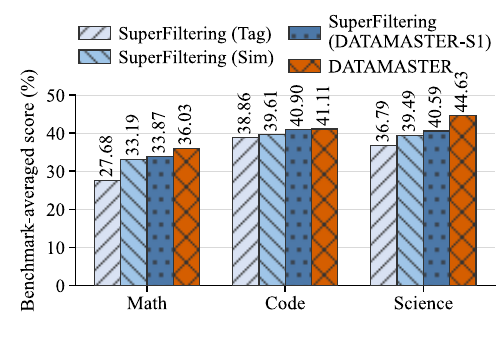}
  \caption{Stage~1 ablation results on OpenHermes-2.5 using Qwen2.5-7B.
  Benchmark-averaged scores are reported for each target domain.
  SuperFiltering (\OurMethod-S1) uses the output of Stage~1 as input
  to SuperFiltering.}
  \label{fig:stage1-ablation}
\end{figure}

To evaluate the effectiveness of the Domain Agent in Stage~1, we integrate Stage~1 as a preprocessing module for the strongest single-dimensional static baseline SuperFiltering. We denote this variant as SuperFiltering (\OurMethod-S1). As shown in \Cref{fig:stage1-ablation}, the experimental results on the three target domains of OpenHermes-2.5 show that SuperFiltering (\OurMethod-S1) achieves consistent improvements compared to other static configurations, thereby demonstrating the clear advantage of dynamic domain selection.
This variant still lags behind the full \OurMethod pipeline in all three experiments, confirming the contribution of Stages~2 to~4 beyond SuperFiltering.


\begin{figure}[t]
  \centering
  \includegraphics[width=\columnwidth]{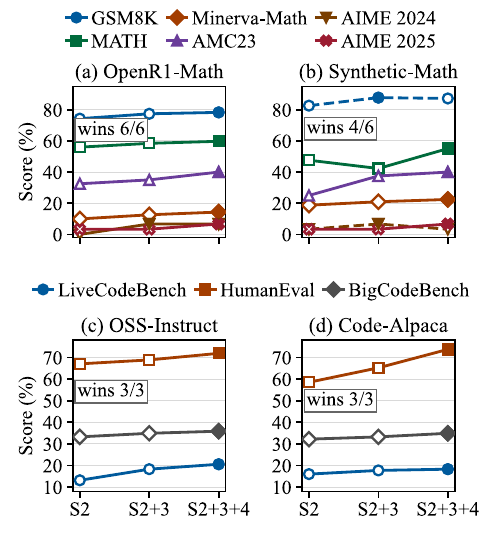}
  \caption{Ablation results for Stages~2 to~4 of \OurMethod.}
  \label{fig:ablation-lines}
\end{figure}

\paragraph{Ablation of Stages~2 to~4.}
For the cascade-depth ablation, we train three configurations per
setting with progressively greater cascade depth. The configurations
comprise \textbf{Stage~2} with the Characteristic Agent only,
\textbf{Stage~2+3} with the Characteristic and Informativeness Agents,
and \textbf{Stage~2+3+4} with the full cascade including the Quality
Agent, as reported in \Cref{tab:main-results}. All other hyperparameters are held fixed. We
evaluate these configurations on four representative Qwen2.5-7B
settings comprising OpenR1-Math, Synthetic-Math, OSS-Instruct, and
Code-Alpaca. Their per-benchmark trajectories appear in
\Cref{fig:ablation-lines}.

The full cascade peaks on $16$ of the $18$ benchmark rows, including
every benchmark in the two code settings. The remaining two rows, both
from Synthetic-Math, reach their highest scores at Stage~2+3. Taken
together, these trajectories support the complementary contributions
of the successive stages across diverse benchmarks.

%% file: Tables/main_results.tex
\begin{table*}[t]
  \centering
  \small
  \begin{tabular}{@{\hspace{3pt}}l@{\hspace{6pt}}l@{\hspace{6pt}}c@{\hspace{3pt}}:@{\hspace{3pt}}c@{\hspace{6pt}}c@{\hspace{6pt}}c@{\hspace{6pt}}c@{\hspace{6pt}}c@{\hspace{6pt}}c@{\hspace{6pt}}c@{\hspace{6pt}}c@{\hspace{3pt}}}
    \toprule
    Target Model & Source pool & \shortstack{Full\\Pool} & Random & \shortstack{Super\\Filtering} & SelectIT & MIG & \shortstack{Select2\\Reason} & DEFT & \shortstack{Claude\\Code} & \OurMethod \\
    \midrule
    \multicolumn{11}{l}{\emph{Math (avg over 6 math benchmarks)}} \\
    Qwen2.5-7B         & OpenR1-Math       & 33.12 & 27.72 & 31.52 & 29.95 & 30.27 & 30.48 & \underline{32.34} & 31.22 & \textbf{34.29} \\
    Qwen2.5-7B         & Synthetic-Math    & 34.58 & 27.33 & 33.59 & 34.30 & 33.75 & 30.54 & \underline{34.68} & 33.23 & \textbf{35.80} \\
    Llama-3.1-8B       & OpenR1-Math       & 20.77 & 16.11 & 18.53 & 17.46 & 20.35 & 18.75 & 19.19 & \textbf{22.41} & \underline{21.21} \\
    Llama-3.1-8B       & Synthetic-Math    & 22.68 & 16.86 & 17.58 & 20.18 & 18.85 & 14.82 & \underline{20.79} & 20.27 & \textbf{21.40} \\
    Olmo-3-7B-Instruct & OpenR1-Math       & 40.80 & 34.25 & 37.02 & 36.99 & 38.78 & 39.74 & 38.96 & \underline{40.47} & \textbf{41.52} \\
    Olmo-3-7B-Instruct & Synthetic-Math    & 40.27 & 37.82 & \underline{41.96} & 39.77 & 39.65 & 39.41 & 41.22 & 41.69 & \textbf{43.07} \\
    \midrule
    \multicolumn{11}{l}{\emph{Medical (avg over 2 medical benchmarks)}} \\
    Qwen2.5-7B         & Medical-Reasoning & 63.97 & 61.08 & 63.38 & 63.55 & 63.79 & 63.11 & \underline{64.52} & 64.42 & \textbf{65.12} \\
    Qwen2.5-7B         & UltraMedical      & 68.03 & 61.64 & 64.03 & 66.24 & 65.07 & 65.16 & 66.69 & \textbf{68.22} & \underline{67.94} \\
    Llama-3.1-8B       & Medical-Reasoning & 61.36 & 52.56 & 60.20 & \underline{62.23} & 58.05 & 60.89 & 57.41 & 62.09 & \textbf{62.50} \\
    Llama-3.1-8B       & UltraMedical      & 67.27 & 61.50 & \textbf{65.91} & 61.44 & 63.60 & 62.23 & 62.74 & 64.22 & \underline{64.97} \\
    Olmo-3-7B-Instruct & Medical-Reasoning & 53.78 & 47.37 & 52.11 & 53.53 & 50.58 & 50.36 & 51.69 & \textbf{53.59} & \underline{53.54} \\
    Olmo-3-7B-Instruct & UltraMedical      & 55.42 & 48.30 & 50.80 & 53.53 & 49.80 & 50.91 & \underline{55.25} & 53.90 & \textbf{55.70} \\
    \midrule
    \multicolumn{11}{l}{\emph{Code (avg over 3 code benchmarks)}} \\
    Qwen2.5-7B         & OSS-Instruct      & 40.60 & 36.39 & 40.44 & \underline{41.42} & 40.96 & 38.60 & 40.75 & 41.40 & \textbf{42.80} \\
    Qwen2.5-7B         & Code-Alpaca       & 38.77 & 35.55 & 38.59 & 36.95 & 38.37 & 38.07 & 40.93 & \underline{40.99} & \textbf{42.33} \\
    Llama-3.1-8B       & OSS-Instruct      & 24.87 & 19.58 & 24.98 & 23.48 & 24.77 & 24.81 & \underline{26.20} & 24.25 & \textbf{26.47} \\
    Llama-3.1-8B       & Code-Alpaca       & 24.14 & 22.31 & \underline{25.36} & 25.06 & 23.77 & 24.90 & 24.02 & 23.34 & \textbf{26.36} \\
    Olmo-3-7B-Instruct & OSS-Instruct      & 42.40 & 35.60 & \underline{40.93} & 39.82 & 36.31 & 37.00 & 37.87 & 39.45 & \textbf{42.27} \\
    Olmo-3-7B-Instruct & Code-Alpaca       & 41.97 & 35.10 & 37.24 & 37.61 & 38.80 & 37.25 & \underline{40.63} & \textbf{43.15} & 40.58 \\
    \bottomrule
  \end{tabular}
  \caption{Single-domain experimental results. Benchmark-averaged scores per
  (target model, source pool) setting. Bold type indicates the row best,
  and \underline{underline} indicates the row second among the seven
  baselines and \OurMethod. Full Pool is excluded from ranking.}
  \label{tab:main-results}
\end{table*}

%% file: Tables/multi_domain.tex
\begin{table*}[t]
  \centering
  \small
  \begin{tabular}{l l c c c c c c}
    \toprule
    \multirow{2}{*}{Source pool} &
    \multirow{2}{*}{Target domain} &
    \multicolumn{2}{c}{SuperFiltering} &
    \multirow{2}{*}{DSIR} &
    \multirow{2}{*}{TSDS} &
    \multirow{2}{*}{Claude Code} &
    \multirow{2}{*}{\OurMethod} \\
    \cmidrule(lr){3-4}
     &  & Tag & Sim &  &  &  &  \\
    \midrule
    OpenHermes-2.5 & Math    & 27.68 & 33.19 & 32.43 & 31.96 & \underline{34.69} & \textbf{36.03} \\
    Tulu-3         & Math    & 22.05 & 25.00 & \underline{30.99} & 28.46 & 30.53 & \textbf{31.32} \\
    OpenHermes-2.5 & Code    & 38.86 & 39.61 & 39.51 & 38.63 & \underline{41.07} & \textbf{41.11} \\
    Tulu-3         & Code    & 37.91 & 40.38 & 39.13 & 39.28 & \underline{40.43} & \textbf{41.65} \\
    OpenHermes-2.5 & Science & 36.79 & 39.49 & 38.55 & 42.63 & \underline{43.59} & \textbf{44.63} \\
    \bottomrule
  \end{tabular}
  \caption{Multi-domain experimental results for Qwen2.5-7B.
  Benchmark-averaged scores per (source pool, target domain) setting.
  Bold type indicates the row best, and \underline{underline} indicates
  the row second among the five baselines and \OurMethod.}
  \label{tab:multi-disc}
\end{table*}

%% file: sections/05-conclusion.tex
\section{Conclusion}
\label{sec:conclusion}

We introduced \OurMethod, an instruction-conditioned agentic data selection system
that translates a user's selection objective into a cascade of
decisions over domain relevance, data characteristics, model-relative
informativeness, and quality. This decomposition is effective because
these criteria play complementary roles, allowing the method to
maintain instruction conditioning throughout selection without
collapsing the objective into a single fixed proxy. Across three target
models and eight source data pools, \OurMethod achieves strong overall
performance against static and agentic baselines in both single- and
multi-domain settings while selecting compact subsets that can surpass
full-pool training. By reducing reliance on manually designed heuristics,
\OurMethod offers a practical path to scalable, task-adaptive data
curation for diverse model development.

\subsubsection{Future Work}
Our evaluated target models are limited to the 7B--8B scale. Future
work will examine whether the observed gains extend to larger and more
architecturally diverse models. The Claude Code comparison uses a
cost-aware prompt and therefore does not constitute an upper bound on
single-pass LLM curation. Future evaluations will broaden this
comparison across curator models, prompts, and inference budgets. A
further improvement to our method is to adapt stage activation and ordering to each
instruction and observed pool while preserving the explicit cascade
structure. Further discussion is provided in Appendix~\ref{app:extensions}.

%% file: sections/appendix.tex

\section{Method Details}
\label{app:info_metric}

\subsection{Negative Log-Likelihood (NLL)}

\begin{equation}
  \mathrm{NLL}_\theta(x, y)
    = -\tfrac{1}{T} \textstyle\sum_{t=1}^{T}
       \log p_\theta(y_t \mid x, y_{<t}).
  \label{eq:nll}
\end{equation}
For a sample $(x,y)$ with reference response
$y=(y_1,\dots,y_T)$, NLL measures the average negative
log-likelihood that the target model $\theta$ assigns to the reference
tokens.

\subsection{Token-Level Entropy}

\begin{equation}
  H_\theta(x, y)
    = \tfrac{1}{T} \textstyle\sum_{t=1}^{T}
       H\!\left(p_\theta(\cdot \mid x, y_{<t})\right).
  \label{eq:entropy}
\end{equation}
Here, $H(\cdot)$ denotes the Shannon entropy of the target model's
next-token distribution. The metric averages this uncertainty over
the reference-response positions.

\subsection{Semantic Drift}

\begin{equation}
  \mathrm{Drift}_\theta(x, y)
    = 1 - \cos\!\bigl(\mathrm{Enc}(\hat{y}),\, \mathrm{Enc}(y)\bigr).
  \label{eq:drift}
\end{equation}
Here, $\hat{y}$ denotes the response generated by the target model for
$x$ before task-specific fine-tuning, and $\mathrm{Enc}$ denotes the
text embedding encoder used to compare the generated and reference
responses. We use \texttt{Qwen/Qwen3-Embedding-0.6B} as the encoder in
our implementation.

\subsection{MeanDiff}

MeanDiff is the per-sample adaptation of the data-parameter resonance
score in ResoFilter~\citep{tu2025resofilter}. For a sample $(x, y)$,
\begin{equation}
  \mathrm{MeanDiff}_\theta(x, y)
    = \tfrac{1}{|\theta_L|} \textstyle\sum_i
      |\theta_L^{(i)} - \tilde{\theta}_L^{(i)}|,
  \label{eq:meandiff}
\end{equation}
where $\theta_L$ is the last-layer up-projection weight and
$\tilde{\theta}_L$ is the same weight after one gradient step on the
sample's loss, with the model restored to $\theta$ before the next
sample. The single-step update uses AdamW with learning rate
$10^{-5}$ on the per-sample loss, and the parameter shift is measured
on the last-layer \texttt{mlp.up\_proj} weight only. ResoFilter
aggregates across the $W_{\mathrm{up}}$ modules of the last $n$
transformer layers, with $n{=}3$ in their default ablation; we restrict
the computation to the last layer to reduce the per-sample
computational cost. The
target model is restored to its pre-step state before the next sample
is processed, so MeanDiff is a per-sample quantity rather than a
cumulative one. We follow ResoFilter's polarity, where low diff corresponds to
in-distribution clean samples and high diff is associated with noisy
or mislabelled samples, by using
$1 - \mathrm{MeanDiff}$ inside the Stage~3 composite
score. A larger MeanDiff weight favors samples that the target model
can absorb without large parameter perturbation, consistent with the
Feature Analysis of \citep{tu2025resofilter}.

\section{Benchmark Details}
\label{app:benchmark-details}

The math benchmarks are GSM8K \citep{cobbe2021gsm8k}, MATH
\citep{hendrycks2021math}, Minerva-Math \citep{lewkowycz2022minerva},
AMC23\footnote{\url{https://maa.org/student-programs/amc/}}, and AIME
2024 and 2025\footnote{\url{https://maa.org/maa-invitational-competitions/}}.
The code benchmarks are LiveCodeBench \citep{jain2024livecodebench},
HumanEval \citep{chen2021humaneval}, and BigCodeBench
\citep{zhuo2024bigcodebench}. The medical benchmarks are MedQA
\citep{jin2021medqa} and MMLU-Medical
\citep{hendrycks2021mmlu}. The science benchmarks used in the
multi-domain experiments are MMLU-STEM \citep{hendrycks2021mmlu} and
GPQA-Diamond \citep{rein2024gpqa}.

\section{Per-Benchmark Results}
\label{app:per-benchmark}

\subsection{Single-Domain Results}
\label{app:per-benchmark-single}

This section reports the per-benchmark scores underlying the
average-score main-results table in the paper. The $18$ (model,
source pool) settings unfold into $66$ (model, source pool, benchmark)
triples: $36$ math, $12$ medical, and $18$ code. Each row reports
the final-epoch score. Within a setting, the seven selection baselines,
\OurMethod, and the \emph{Full Pool} control all use the setting's
single fixed LR, matching the paper's experimental protocol. The
selection methods use the same 10K subset budget, whereas
\emph{Full Pool} uses the entire uncurated source pool and is excluded
from the ranking. Setting-level averages are computed from the
underlying unrounded benchmark scores and then rounded to two decimal
places; the per-benchmark scores shown here are also rounded to two
decimal places. Bold marks the row best and underline the row second
among the eight selection methods.

\Cref{tab:per-benchmark-math,tab:per-benchmark-medical,tab:per-benchmark-code}
together support the setting-level claim in the paper's single-domain experiments.
Measured against the six static baselines,
\OurMethod is the row best on $41$ of the $66$ benchmark rows.
Adding the dynamic Claude Code curator into the ranking changes
\OurMethod's per-benchmark top-1 count to $33$ of $66$ and its top-2
count to $53$ of $66$, with Claude Code being the most frequent alternative
leader. These per-benchmark results complement the setting-level
comparison in the main paper, where \OurMethod achieves a higher average
score than Claude Code in $14$ of $18$ settings.

\begin{table*}[!t]
  \centering
  \small
  \setlength{\tabcolsep}{3pt}
  \begin{tabular}{@{}l l l c c c c c c c c c@{}}
    \toprule
    Model & Source pool & Benchmark & Rand & SupFilt & Sel2Rsn & MIG & SelIT & DEFT & \shortstack{Claude\\Code} & \OurMethod & \shortstack{Full\\Pool} \\
    \midrule
    Qwen2.5-7B & OpenR1-Math & GSM8K & 69.85 & 75.74 & 72.76 & 71.87 & 73.01 & \underline{76.04} & 73.46 & \textbf{78.32} & 76.12 \\
     &  & MATH & 53.64 & \underline{56.98} & 54.93 & 55.24 & 56.12 & 56.56 & 56.04 & \textbf{59.76} & 58.52 \\
     &  & Minerva-Math & 10.34 & 10.60 & 11.03 & 12.87 & 12.24 & 13.95 & \textbf{16.18} & \underline{14.34} & 13.24 \\
     &  & AMC23 & 32.50 & \textbf{42.50} & 37.50 & 35.00 & 35.00 & 37.50 & 35.00 & \underline{40.00} & 37.50 \\
     &  & AIME24 & 0.00 & 0.00 & \textbf{6.67} & \underline{3.33} & \underline{3.33} & \underline{3.33} & \underline{3.33} & \textbf{6.67} & 10.00 \\
     &  & AIME25 & 0.00 & \underline{3.33} & 0.00 & \underline{3.33} & 0.00 & \textbf{6.67} & \underline{3.33} & \textbf{6.67} & 3.33 \\
    \cmidrule(lr){1-12}
    Qwen2.5-7B & Synthetic-Math & GSM8K & 69.22 & 83.17 & 78.32 & 87.11 & \textbf{87.87} & 86.96 & \underline{87.57} & 87.26 & 87.04 \\
     &  & MATH & 48.52 & 52.38 & 50.80 & 54.32 & \textbf{55.94} & 52.68 & 54.68 & \underline{55.08} & 56.24 \\
     &  & Minerva-Math & 18.75 & \textbf{29.32} & 19.12 & 20.22 & 19.49 & 20.96 & 21.32 & \underline{22.43} & 21.69 \\
     &  & AMC23 & 27.50 & 30.00 & 35.00 & \underline{37.50} & 32.50 & \underline{37.50} & 32.50 & \textbf{40.00} & 32.50 \\
     &  & AIME24 & 0.00 & \underline{3.33} & 0.00 & 0.00 & \textbf{6.67} & \textbf{6.67} & 0.00 & \underline{3.33} & 6.67 \\
     &  & AIME25 & 0.00 & \underline{3.33} & 0.00 & \underline{3.33} & \underline{3.33} & \underline{3.33} & \underline{3.33} & \textbf{6.67} & 3.33 \\
    \cmidrule(lr){1-12}
    Llama-3.1-8B & OpenR1-Math & GSM8K & 57.47 & 65.68 & 70.66 & 70.53 & 67.82 & 62.09 & \underline{71.07} & \textbf{74.60} & 63.61 \\
     &  & MATH & 19.28 & 21.22 & 21.53 & 18.88 & 16.46 & \textbf{26.84} & \underline{25.44} & 23.62 & 30.34 \\
     &  & Minerva-Math & 9.93 & 11.76 & 10.29 & 11.03 & 10.50 & 12.87 & \textbf{13.79} & \underline{13.24} & 10.66 \\
     &  & AMC23 & 10.00 & 12.50 & 10.00 & \underline{15.00} & 10.00 & 10.00 & \textbf{17.50} & 12.50 & 20.00 \\
     &  & AIME24 & \underline{0.00} & \underline{0.00} & \underline{0.00} & \textbf{3.33} & \underline{0.00} & \textbf{3.33} & \textbf{3.33} & \textbf{3.33} & 0.00 \\
     &  & AIME25 & \underline{0.00} & \underline{0.00} & \underline{0.00} & \textbf{3.33} & \underline{0.00} & \underline{0.00} & \textbf{3.33} & \underline{0.00} & 0.00 \\
    \cmidrule(lr){1-12}
    Llama-3.1-8B & Synthetic-Math & GSM8K & 63.91 & 62.93 & 51.25 & 70.28 & \textbf{73.74} & \underline{73.09} & 68.84 & 72.86 & 76.65 \\
     &  & MATH & 17.30 & 17.74 & 15.84 & 20.10 & 19.78 & 20.26 & \underline{21.14} & \textbf{21.96} & 24.90 \\
     &  & Minerva-Math & 12.44 & 14.82 & 14.34 & 15.19 & 15.07 & 14.71 & \underline{15.81} & \textbf{16.91} & 16.18 \\
     &  & AMC23 & 7.50 & \underline{10.00} & 7.50 & 7.50 & \textbf{12.50} & \underline{10.00} & \textbf{12.50} & \underline{10.00} & 15.00 \\
     &  & AIME24 & \underline{0.00} & \underline{0.00} & \underline{0.00} & \underline{0.00} & \underline{0.00} & \textbf{3.33} & \underline{0.00} & \textbf{3.33} & 0.00 \\
     &  & AIME25 & \underline{0.00} & \underline{0.00} & \underline{0.00} & \underline{0.00} & \underline{0.00} & \textbf{3.33} & \textbf{3.33} & \textbf{3.33} & 3.33 \\
    \cmidrule(lr){1-12}
    Olmo-3-7B-Instruct & OpenR1-Math & GSM8K & 83.78 & 86.50 & 90.75 & 90.27 & 90.59 & 91.96 & \underline{92.27} & \textbf{92.80} & 92.65 \\
     &  & MATH & 59.80 & 64.24 & 69.42 & 63.02 & 65.06 & 69.20 & \textbf{71.40} & \underline{69.48} & 70.52 \\
     &  & Minerva-Math & 16.91 & 23.90 & 24.10 & \underline{25.24} & 24.63 & 24.26 & 25.00 & \textbf{26.84} & 22.43 \\
     &  & AMC23 & 35.00 & \underline{37.50} & \underline{37.50} & \underline{37.50} & 35.00 & 35.00 & \underline{37.50} & \textbf{40.00} & 42.50 \\
     &  & AIME24 & 6.67 & 6.67 & \textbf{13.33} & \underline{10.00} & 3.33 & 6.67 & \underline{10.00} & \underline{10.00} & 10.00 \\
     &  & AIME25 & 3.33 & 3.33 & 3.33 & \underline{6.67} & 3.33 & \underline{6.67} & \underline{6.67} & \textbf{10.00} & 6.67 \\
    \cmidrule(lr){1-12}
    Olmo-3-7B-Instruct & Synthetic-Math & GSM8K & 86.76 & 90.07 & 89.99 & 88.42 & 89.15 & 90.07 & \textbf{91.96} & \underline{90.98} & 91.28 \\
     &  & MATH & 65.94 & \textbf{71.74} & 67.84 & 69.82 & 68.62 & 71.10 & \underline{71.58} & 71.24 & 71.44 \\
     &  & Minerva-Math & 20.90 & 22.43 & \underline{22.79} & 21.32 & 21.69 & 21.98 & \textbf{24.93} & 22.06 & 23.90 \\
     &  & AMC23 & 40.00 & \textbf{47.50} & 42.50 & \underline{45.00} & 42.50 & \textbf{47.50} & \underline{45.00} & \textbf{47.50} & 45.00 \\
     &  & AIME24 & 6.67 & \textbf{13.33} & 3.33 & 6.67 & 6.67 & 6.67 & \underline{10.00} & \textbf{13.33} & 3.33 \\
     &  & AIME25 & 6.67 & 6.67 & \underline{10.00} & 6.67 & \underline{10.00} & \underline{10.00} & 6.67 & \textbf{13.33} & 6.67 \\
    \bottomrule
  \end{tabular}
  \caption{Per-benchmark scores: math domain. Final-epoch
  score per row. Within each setting, all result columns use the same
  fixed LR from
  $\{1\!\times\!10^{-5},5\!\times\!10^{-6},1\!\times\!10^{-6}\}$.
  The eight selection methods use the same 10K budget, while
  Full Pool trains on the full uncurated pool and is not
  ranked. Bold type indicates the row best and \underline{underline} = row
  second among the eight selection methods.}
  \label{tab:per-benchmark-math}
\end{table*}

\begin{table*}[!t]
  \centering
  \small
  \setlength{\tabcolsep}{3pt}
  \begin{tabular}{@{}l l l c c c c c c c c c@{}}
    \toprule
    Model & Source pool & Benchmark & Rand & SupFilt & Sel2Rsn & MIG & SelIT & DEFT & \shortstack{Claude\\Code} & \OurMethod & \shortstack{Full\\Pool} \\
    \midrule
    Qwen2.5-7B & Medical-Reasoning & MedQA & 59.39 & 61.51 & 61.90 & \underline{62.37} & 61.67 & 62.06 & 62.14 & \textbf{63.55} & 59.62 \\
     &  & MMLU-Medical & 62.76 & 65.25 & 64.33 & 65.21 & 65.42 & \textbf{66.98} & \underline{66.70} & \underline{66.70} & 68.33 \\
    \cmidrule(lr){1-12}
    Qwen2.5-7B & UltraMedical & MedQA & 58.51 & 60.64 & 62.84 & 63.94 & 62.92 & 63.71 & \underline{65.04} & \textbf{65.36} & 67.56 \\
     &  & MMLU-Medical & 64.78 & 67.43 & 67.49 & 66.20 & 69.56 & 69.67 & \textbf{71.41} & \underline{70.52} & 68.50 \\
    \cmidrule(lr){1-12}
    Llama-3.1-8B & Medical-Reasoning & MedQA & 51.14 & \textbf{62.22} & 58.44 & 59.70 & \underline{61.67} & 58.05 & 59.81 & 60.09 & 58.92 \\
     &  & MMLU-Medical & 53.98 & 58.18 & 63.35 & 56.39 & 62.78 & 56.78 & \underline{64.37} & \textbf{64.91} & 63.79 \\
    \cmidrule(lr){1-12}
    Llama-3.1-8B & UltraMedical & MedQA & 58.60 & \textbf{63.16} & 60.33 & 61.67 & 59.31 & 60.41 & 61.19 & \underline{62.29} & 64.81 \\
     &  & MMLU-Medical & 64.41 & \textbf{68.67} & 64.13 & 65.53 & 63.57 & 65.08 & 67.26 & \underline{67.66} & 69.73 \\
    \cmidrule(lr){1-12}
    Olmo-3-7B-Instruct & Medical-Reasoning & MedQA & 40.23 & 43.28 & 44.38 & 44.15 & 45.46 & 45.48 & \underline{45.91} & \textbf{46.43} & 48.21 \\
     &  & MMLU-Medical & 54.51 & 60.93 & 56.33 & 57.01 & \textbf{61.60} & 57.90 & \underline{61.27} & 60.65 & 59.36 \\
    \cmidrule(lr){1-12}
    Olmo-3-7B-Instruct & UltraMedical & MedQA & 40.65 & 41.79 & 45.40 & 42.81 & 49.33 & \underline{49.80} & 47.82 & \textbf{50.04} & 51.37 \\
     &  & MMLU-Medical & 55.94 & 59.81 & 56.42 & 56.78 & 57.74 & \underline{60.71} & 59.98 & \textbf{61.35} & 59.47 \\
    \bottomrule
  \end{tabular}
  \caption{Per-benchmark scores: medical domain. Layout follows \Cref{tab:per-benchmark-math}.}
  \label{tab:per-benchmark-medical}
\end{table*}

\begin{table*}[!t]
  \centering
  \small
  \setlength{\tabcolsep}{3pt}
  \begin{tabular}{@{}l l l c c c c c c c c c@{}}
    \toprule
    Model & Source pool & Benchmark & Rand & SupFilt & Sel2Rsn & MIG & SelIT & DEFT & \shortstack{Claude\\Code} & \OurMethod & \shortstack{Full\\Pool} \\
    \midrule
    Qwen2.5-7B & OSS-Instruct & LiveCodeBench & 15.08 & 16.57 & 17.71 & 17.14 & 18.29 & 17.14 & \underline{19.43} & \textbf{20.57} & 18.86 \\
     &  & HumanEval & 62.80 & 68.95 & 63.41 & 69.51 & 69.82 & \underline{70.29} & 69.51 & \textbf{71.95} & 68.90 \\
     &  & BigCodeBench & 31.28 & 35.79 & 34.69 & \textbf{36.23} & \underline{36.14} & 34.82 & 35.26 & 35.88 & 34.04 \\
    \cmidrule(lr){1-12}
    Qwen2.5-7B & Code-Alpaca & LiveCodeBench & 14.29 & 14.86 & 16.00 & 16.57 & 15.43 & 18.02 & \textbf{18.54} & \underline{18.29} & 17.14 \\
     &  & HumanEval & 64.63 & 70.12 & 67.68 & 66.46 & 64.02 & \underline{71.95} & 68.90 & \textbf{73.78} & 66.46 \\
     &  & BigCodeBench & 27.72 & 30.79 & 30.53 & 32.09 & 31.40 & 32.81 & \textbf{35.53} & \underline{34.91} & 32.72 \\
    \cmidrule(lr){1-12}
    Llama-3.1-8B & OSS-Instruct & LiveCodeBench & 1.71 & 3.43 & 2.29 & \underline{4.00} & 2.86 & \textbf{4.12} & \underline{4.00} & 3.43 & 4.57 \\
     &  & HumanEval & 37.20 & 45.56 & 45.83 & 43.29 & 44.95 & \underline{47.56} & 42.07 & \textbf{48.78} & 44.51 \\
     &  & BigCodeBench & 19.82 & 25.96 & 26.32 & \underline{27.02} & 22.63 & 26.93 & 26.67 & \textbf{27.19} & 25.53 \\
    \cmidrule(lr){1-12}
    Llama-3.1-8B & Code-Alpaca & LiveCodeBench & 8.00 & \textbf{12.00} & 10.35 & 9.71 & \underline{11.43} & \textbf{12.00} & 9.14 & 10.86 & 8.57 \\
     &  & HumanEval & 38.75 & 41.46 & \textbf{43.29} & 39.32 & 40.07 & 39.63 & 38.41 & \underline{42.68} & 40.24 \\
     &  & BigCodeBench & 20.18 & 22.62 & 21.05 & 22.28 & \underline{23.68} & 20.42 & 22.46 & \textbf{25.53} & 23.60 \\
    \cmidrule(lr){1-12}
    Olmo-3-7B-Instruct & OSS-Instruct & LiveCodeBench & 17.48 & 19.92 & 19.43 & 19.26 & \textbf{21.71} & 19.43 & 20.57 & \underline{20.92} & 21.14 \\
     &  & HumanEval & 61.76 & \underline{72.61} & 62.80 & 60.37 & 70.29 & 64.63 & 67.07 & \textbf{75.00} & 74.92 \\
     &  & BigCodeBench & 27.57 & 30.26 & 28.77 & 29.30 & 27.46 & 29.56 & \underline{30.70} & \textbf{30.88} & 31.14 \\
    \cmidrule(lr){1-12}
    Olmo-3-7B-Instruct & Code-Alpaca & LiveCodeBench & 16.00 & 14.86 & 17.14 & 18.04 & 19.57 & 20.57 & \textbf{22.29} & \underline{21.71} & 20.57 \\
     &  & HumanEval & 65.24 & 70.73 & 68.76 & \underline{73.17} & 69.51 & \underline{73.17} & \textbf{78.66} & 72.56 & 76.22 \\
     &  & BigCodeBench & 24.05 & 26.14 & 25.84 & 25.19 & 23.75 & \underline{28.16} & \textbf{28.51} & 27.46 & 29.12 \\
    \bottomrule
  \end{tabular}
  \caption{Per-benchmark scores: code domain. Layout follows \Cref{tab:per-benchmark-math}.}
  \label{tab:per-benchmark-code}
\end{table*}

\subsection{Multi-Domain Experiment Results}
\label{app:per-benchmark-multi}

This section reports the per-benchmark scores underlying the
multi-domain average-score table and Stage~1 selection ablation in the
paper. All runs fine-tune Qwen2.5-7B and share each setting's single
fixed learning rate, matching the protocol of the paper's multi-domain
experiments. \Cref{tab:multidomain-per-benchmark} unfolds the five
$(\text{source pool},\text{target domain})$ settings into $20$
benchmark rows: $6$ math, $3$ code, and $2$ science rows for the three
OpenHermes-2.5 targets, plus $6$ math and $3$ code rows for the two
Tulu-3 targets.
\Cref{tab:stage1-per-benchmark} separately reports the selection
ablation for Stage~1 on the three OpenHermes-2.5 target domains.
SuperFiltering (\OurMethod-S1) applies SuperFiltering after
\OurMethod's Stage~1 cluster matching. \textbf{Bold} marks the row best
and \underline{underline} the row second among the methods in each
table.

The math, code, and science target domains use the benchmark suites
documented in Appendix~\ref{app:benchmark-details}.

\begin{table*}[!t]
  \centering
  \small
  \setlength{\tabcolsep}{4pt}
  \begin{tabular}{l l l c c c c c c}
    \toprule
    Source pool & \shortstack{Target\\domain} & Benchmark & \multicolumn{2}{c}{SuperFiltering} & DSIR & TSDS & \shortstack{Claude\\Code} & \OurMethod \\
    \cmidrule(lr){4-5}
     &  &  & Tag & Sim &  &  &  &  \\
    \midrule
    OpenHermes-2.5 & Math & GSM8K & 50.87 & \underline{85.29} & 81.71 & 72.48 & 84.76 & \textbf{86.73} \\
     &  & MATH & 53.12 & 52.22 & \underline{54.94} & 54.02 & \textbf{55.24} & 54.34 \\
     &  & Minerva-Math & 15.44 & \underline{19.12} & 15.44 & 16.91 & 17.28 & \textbf{24.26} \\
     &  & AMC23 & \textbf{40.00} & 32.50 & 32.50 & 35.00 & \underline{37.50} & \underline{37.50} \\
     &  & AIME24 & 3.33 & 3.33 & \underline{6.67} & \textbf{10.00} & \underline{6.67} & \underline{6.67} \\
     &  & AIME25 & \underline{3.33} & \textbf{6.67} & \underline{3.33} & \underline{3.33} & \textbf{6.67} & \textbf{6.67} \\
    \cmidrule(lr){1-9}
    Tulu-3 & Math & GSM8K & 55.80 & 58.07 & \underline{80.36} & 76.57 & 75.74 & \textbf{84.46} \\
     &  & MATH & 39.10 & 41.96 & \underline{46.94} & 44.24 & 46.76 & \textbf{47.06} \\
     &  & Minerva-Math & 16.54 & 19.12 & \underline{19.49} & 15.81 & 16.54 & \textbf{20.59} \\
     &  & AMC23 & 17.50 & 27.50 & \underline{32.50} & 27.50 & \textbf{37.50} & \underline{32.50} \\
     &  & AIME24 & \underline{3.33} & 0.00 & \underline{3.33} & \underline{3.33} & \textbf{6.67} & \underline{3.33} \\
     &  & AIME25 & \underline{0.00} & \textbf{3.33} & \textbf{3.33} & \textbf{3.33} & \underline{0.00} & \underline{0.00} \\
    \cmidrule(lr){1-9}
    OpenHermes-2.5 & Code & LiveCodeBench & 16.00 & 14.86 & 15.43 & \textbf{18.29} & \underline{17.29} & 17.14 \\
     &  & HumanEval & 67.68 & \underline{69.51} & 67.93 & 66.29 & \textbf{70.12} & \underline{69.51} \\
     &  & BigCodeBench & 32.89 & 34.47 & 35.18 & 31.32 & \underline{35.79} & \textbf{36.67} \\
    \cmidrule(lr){1-9}
    Tulu-3 & Code & LiveCodeBench & 17.14 & 18.29 & 16.57 & 16.04 & \underline{18.63} & \textbf{18.86} \\
     &  & HumanEval & 62.20 & \underline{67.68} & 63.29 & 66.80 & 67.39 & \textbf{69.51} \\
     &  & BigCodeBench & 34.39 & 35.18 & \textbf{37.54} & 35.00 & 35.26 & \underline{36.58} \\
    \cmidrule(lr){1-9}
    OpenHermes-2.5 & Science & GPQA-Diamond & 23.23 & 22.22 & 23.74 & 26.26 & \underline{27.27} & \textbf{28.79} \\
     &  & MMLU-STEM & 50.35 & 56.76 & 53.35 & 59.00 & \underline{59.91} & \textbf{60.47} \\
    \bottomrule
  \end{tabular}
  \caption{Per-benchmark multi-domain scores (Qwen2.5-7B).
  Final-epoch score per row; all reported methods in a setting share
  that setting's single fixed LR under the main-paper protocol.
  SuperFiltering (Tag), SuperFiltering (Sim),
  DSIR, TSDS, and Claude Code are the baselines
  evaluated in the paper's multi-domain experiments.
  Bold type indicates the row best,
  \underline{underline} = row second among the reported methods. Math averages the
  six math benchmarks, Code the three code benchmarks, and Science the
  two science benchmarks (MMLU-STEM and GPQA-Diamond).}
  \label{tab:multidomain-per-benchmark}
\end{table*}

\begin{table*}[!t]
  \centering
  \small
  \setlength{\tabcolsep}{6pt}
  \begin{tabular}{l l c c c c}
    \toprule
    \shortstack{Target\\domain} & Benchmark
      & \multicolumn{2}{c}{SuperFiltering}
      & \shortstack{SuperFiltering\\(\OurMethod-S1)}
      & \OurMethod \\
    \cmidrule(lr){3-4}
     &  & Tag & Sim &  &  \\
    \midrule
    Math & GSM8K & 50.87 & \underline{85.29} & 83.62 & \textbf{86.73} \\
     & MATH & 53.12 & 52.22 & \underline{54.00} & \textbf{54.34} \\
     & Minerva-Math & 15.44 & \underline{19.12} & 17.28 & \textbf{24.26} \\
     & AMC23 & \textbf{40.00} & 32.50 & 35.00 & \underline{37.50} \\
     & AIME24 & \underline{3.33} & \underline{3.33} & \textbf{6.67} & \textbf{6.67} \\
     & AIME25 & \underline{3.33} & \textbf{6.67} & \textbf{6.67} & \textbf{6.67} \\
    \cmidrule(lr){1-6}
    Code & LiveCodeBench & 16.00 & 14.86 & \textbf{18.86} & \underline{17.14} \\
     & HumanEval & 67.68 & \textbf{69.51} & \underline{68.49} & \textbf{69.51} \\
     & BigCodeBench & 32.89 & 34.47 & \underline{35.35} & \textbf{36.67} \\
    \cmidrule(lr){1-6}
    Science & GPQA-Diamond & 23.23 & 22.22 & \underline{24.95} & \textbf{28.79} \\
     & MMLU-STEM & 50.35 & \underline{56.76} & 56.22 & \textbf{60.47} \\
    \bottomrule
  \end{tabular}
  \caption{Per-benchmark Stage~1 selection ablation on
  OpenHermes-2.5 using Qwen2.5-7B. SuperFiltering
  (\OurMethod-S1) applies SuperFiltering to the pool selected by
  \OurMethod's Stage~1 Domain Agent. Final-epoch scores are reported
  under the same fixed learning rate used by all methods in each
  target-domain setting. Bold type indicates the row best and
  \underline{underline} the row second among the four reported
  configurations.}
  \label{tab:stage1-per-benchmark}
\end{table*}

\section{Per-Domain Task Instructions}
\label{app:task-instructions}

\Cref{tab:task-instructions} gives representative natural-language
task instructions $\mathcal{I}$ for the three single-domain settings, and
\Cref{tab:multidomain-instructions} gives representative
target-selection instructions for the paper's multi-domain
comparison. The two settings differ in
intent: the former curates a high-quality subset from a pool that is
already on-domain, whereas the latter must first carve a single target
domain out of a genuinely mixed pool. In both cases the same
instruction is read by every stage's agent, which infers its per-stage
configuration from this text together with its inspection of the pool.

\definecolor{mathrow}{HTML}{E8F0FE}
\definecolor{coderow}{HTML}{E6F4EA}
\definecolor{medrow}{HTML}{FCE8E6}
\definecolor{scirow}{HTML}{ECE4F6}

\begin{table*}[!t]
  \centering
  \small
  \setlength{\tabcolsep}{5pt}
  \begin{tabular}{@{}p{1.9cm} p{\dimexpr\textwidth-3.0cm\relax}@{}}
    \toprule
    Domain & Task instruction $\mathcal{I}$ \\
    \midrule
    \rowcolor{mathrow}
    \textbf{Math}\newline\emph{verifiable rigour} &
    Curate a math reasoning subset from this single-domain pool. Keep
    correct, rigorously step-by-step solutions with verifiable final
    answers, preserve LaTeX, and drop unverified or sloppily reasoned
    entries. \\
    \rowcolor{coderow}
    \textbf{Code}\newline\emph{functional correctness} &
    Curate a coding subset from this single-domain programming pool.
    Keep correct, runnable, cleanly structured solutions, reward
    non-trivial coding patterns, and discard buggy or unverified
    snippets. \\
    \rowcolor{medrow}
    \textbf{Medical}\newline\emph{factual, safe reasoning} &
    Curate a medical reasoning subset from this single-domain clinical
    pool. Keep factually accurate answers with sound multi-step
    clinical reasoning, retain precise terminology, and drop unreliable
    or unsafe responses. \\
    \bottomrule
  \end{tabular}
  \caption{Single-domain task instructions. Representative
  natural-language instructions $\mathcal{I}$ for \OurMethod when the source
  pool is already on-domain. Row colour marks the
  domain. The multi-domain selection instructions are separate, in
  \Cref{tab:multidomain-instructions}.}
  \label{tab:task-instructions}
\end{table*}

\begin{table*}[!t]
  \centering
  \small
  \setlength{\tabcolsep}{5pt}
  \begin{tabular}{@{}p{1.9cm} p{\dimexpr\textwidth-3.0cm\relax}@{}}
    \toprule
    Target domain & Selection instruction $\mathcal{I}$ \\
    \midrule
    \rowcolor{mathrow}
    \textbf{Math}\newline\emph{select + verify} &
    From this multi-domain instruction pool, identify and select the
    mathematics samples, keeping problems with correct, rigorously
    step-by-step solutions and verifiable final answers while
    preserving mathematical notation. \\
    \rowcolor{coderow}
    \textbf{Code}\newline\emph{select + runnable} &
    From this multi-domain instruction pool, select the programming
    samples, retaining tasks whose code is correct, runnable, and
    cleanly structured, and discarding buggy or trivial snippets. \\
    \rowcolor{scirow}
    \textbf{Science}\newline\emph{select + factual} &
    From this multi-domain instruction pool, select the natural-science
    samples spanning physics, chemistry, and biology, favoring
    factually accurate explanations grounded in sound scientific
    reasoning. \\
    \bottomrule
  \end{tabular}
  \caption{Multi-domain selection instructions.
  Representative instructions $\mathcal{I}$ for \OurMethod when the source
  is a mixed pool and the target is one domain. Unlike the
  single-domain curation
  instructions of \Cref{tab:task-instructions}, each one must first
  select its target domain from the mixture. Row colour marks the
  target domain.}
  \label{tab:multidomain-instructions}
\end{table*}

\section{Representative Informativeness-Agent Weight Allocations}
\label{app:inf-weights}


\Cref{tab:inf-weights} presents representative metric-weight vectors
$[w^1, \dots, w^{m_1}]$ emitted by the Informativeness Agent for the task
instructions and observed data characteristics in our evaluated domains
(math, code, medical, and science). These examples are generated by
the $P_{inf}$ decision (\Cref{tab:prompt-inf}), which reads the
corresponding instruction from \Cref{tab:task-instructions} or
\Cref{tab:multidomain-instructions} together with tool-assisted
observations of the corresponding data pool.
They illustrate the agent's context-dependent allocation behavior;
the weights are generated at run time rather than treated as fixed
hyperparameters.
The implementation key \texttt{loss} in the prompt and table denotes
the NLL metric defined in Appendix~A.

In these representative outputs, the allocations track the
instruction cues that $P_{inf}$
associates with each metric, and the agent spreads weight over all
four metrics, expressing low relevance as a small share (down to
$0.05$) rather than exclusion. On every math row \texttt{drift}
carries the largest (or tied-largest, on Synthetic-Math) share
($0.35$--$0.40$) with \texttt{mean\_diff} the other top share
($0.25$--$0.35$), matching the instructions' emphasis on
\emph{rigorously step-by-step} solutions (a desired output form)
and \emph{verifiable} correctness (clean, reliable samples). On
every code row \texttt{mean\_diff} is the largest share
($0.35$--$0.45$) and \texttt{loss} the second
($0.25$--$0.35$), matching the cues toward \emph{runnable, cleanly
structured} solutions and \emph{non-trivial coding patterns}, while
\texttt{drift} and \texttt{entropy} split the small remainder. On
the medical pools the agent again favors \texttt{drift}
($0.35$--$0.40$) and \texttt{mean\_diff} ($0.30$--$0.35$): it reads
\emph{multi-step clinical reasoning} as the binding output-form
requirement and factual reliability as the secondary one, while the
knowledge-oriented cue \emph{precise terminology} yields a moderate
\texttt{loss} share ($0.15$--$0.20$). The science selection
instead puts its weight on the knowledge side: the largest
\texttt{loss} share of its row ($0.30$) plus the largest
\texttt{entropy} share of any row ($0.25$), consistent with
knowledge coverage rather than output form being the binding
requirement behind \emph{factually accurate explanations}. Two
regularities illustrate the intended context-sensitive behavior: the
dominant metric of each domain persists from the single-domain
curation rows to the multi-domain selection rows (\texttt{drift}
for math, \texttt{mean\_diff} for code), and the two single-domain
pools of a domain never differ by more than $0.10$ on any metric,
consistent with the instruction cues providing the main allocation
signal while the observed data characteristics adjust it.

\begin{table*}[!t]
  \centering
  \small
  \setlength{\tabcolsep}{5pt}
  \begin{tabular}{@{}l l c c c c@{}}
    \toprule
    Setting & Source pool
    & \texttt{loss} & \texttt{entropy} & \texttt{drift} & \texttt{mean\_diff} \\
    \midrule
    \multicolumn{6}{@{}p{9.2cm}@{}}{\emph{Math.} Cues:
      ``rigorously step-by-step'' $\rightarrow$ \texttt{drift};
      ``correct \dots\ verifiable final answers'' $\rightarrow$
      \texttt{mean\_diff}.} \\[2pt]
    single & OpenR1-Math    & 0.05 & 0.20 & 0.40 & 0.35 \\
    single & Synthetic-Math & 0.15 & 0.15 & 0.35 & 0.35 \\
    multi  & OpenHermes-2.5 & 0.25 & 0.15 & 0.35 & 0.25 \\
    multi  & Tulu-3         & 0.15 & 0.15 & 0.40 & 0.30 \\
    \midrule
    \multicolumn{6}{@{}p{9.2cm}@{}}{\emph{Code.} Cues:
      ``correct, runnable, cleanly structured'' $\rightarrow$
      \texttt{mean\_diff}; ``non-trivial coding patterns''
      $\rightarrow$ \texttt{loss}.} \\[2pt]
    single & OSS-Instruct   & 0.25 & 0.20 & 0.15 & 0.40 \\
    single & Code-Alpaca    & 0.25 & 0.20 & 0.20 & 0.35 \\
    multi  & OpenHermes-2.5 & 0.30 & 0.15 & 0.20 & 0.35 \\
    multi  & Tulu-3         & 0.35 & 0.10 & 0.10 & 0.45 \\
    \midrule
    \multicolumn{6}{@{}p{9.2cm}@{}}{\emph{Medical.} Cues:
      ``multi-step clinical reasoning'' $\rightarrow$ \texttt{drift};
      ``factually accurate \dots\ drop unreliable'' $\rightarrow$
      \texttt{mean\_diff}; ``precise terminology'' (long-tail
      knowledge) $\rightarrow$ \texttt{loss}.} \\[2pt]
    single & Medical-Reasoning & 0.20 & 0.15 & 0.35 & 0.30 \\
    single & UltraMedical      & 0.15 & 0.10 & 0.40 & 0.35 \\
    \midrule
    \multicolumn{6}{@{}p{9.2cm}@{}}{\emph{Science.} Cues:
      knowledge-heavy target $\rightarrow$ \texttt{loss},
      \texttt{entropy}; ``factually accurate explanations''
      $\rightarrow$ \texttt{mean\_diff}.} \\[2pt]
    multi  & OpenHermes-2.5 & 0.30 & 0.25 & 0.25 & 0.20 \\
    \bottomrule
  \end{tabular}
  \caption{Representative Informativeness-Agent weight
  allocations. Example four-metric vectors emitted under $P_{inf}$
  (\Cref{tab:prompt-inf}) for the instructions and tool-assisted
  observations in the evaluated domains. \emph{single} rows use the
  single-domain curation instructions of \Cref{tab:task-instructions}
  and \emph{multi} rows the selection instructions of
  \Cref{tab:multidomain-instructions}. The cue $\rightarrow$ metric
  mapping in each group header explains the dominant weights. These
  representative outputs illustrate context-dependent allocation;
  they are not fixed method hyperparameters.}
  \label{tab:inf-weights}
\end{table*}

\section{Data Pool Sources and Licenses}
\label{app:data-licenses}

\paragraph{Data pool selection rationale.}
The single-domain study uses two public SFT pools per domain that differ
in scale and construction. The math pools are OpenR1-Math
(94K)\footnote{\url{https://huggingface.co/datasets/open-r1/OpenR1-Math-220k}}
and Synthetic-Math
(100K)\footnote{\url{https://www.primeintellect.ai/blog/synthetic-1-release}};
the medical pools are the English split of Medical-Reasoning
(20K) \citep{chen2024huatuogpto1} and UltraMedical (410K)
\citep{zhang2024ultramedical}; and the code pools are OSS-Instruct
(50K) \citep{wei2024selfcodealign} and Code-Alpaca
(20K)\footnote{\url{https://github.com/sahil280114/codealpaca}}.
This design allows source-pool
robustness to be examined while the target domain is held fixed.
The multi-domain study instead uses OpenHermes-2.5
(1M)\footnote{\url{https://huggingface.co/datasets/teknium/OpenHermes-2.5}}
and Tulu-3-SFT-Mixture (939K)
\citep{lambert2024tulu3}, two large public mixed-domain SFT pools with
overlapping math and code content but different compositions. This design
allows target-domain selection to be compared across source mixtures under
the same 10K budget. Science is evaluated only on OpenHermes-2.5 because
Tulu-3-SFT-Mixture does not provide a coherent science subset. The benchmark
suite complements these source-pool comparisons with open-ended reasoning,
executable code generation, and domain-specific multiple-choice evaluation.

\paragraph{Dataset licenses.}
\Cref{tab:licenses} lists the license of every source pool used for
curation. All are used for research purposes only, consistent with
their terms. Two carry no single permissive license: OpenHermes-2.5
mixes many sources (some derived from proprietary model outputs) and
states no single license, and Tulu-3-SFT-Mixture is ODC-BY-1.0 overall
but contains non-commercial subsets; we treat both as research-use.

\begin{table}[!t]
  \centering
  \small
  \setlength{\tabcolsep}{4pt}
  \begin{tabular}{@{}ll@{}}
    \toprule
    Source pool & License \\
    \midrule
    OpenR1-Math        & Apache-2.0 \\
    Synthetic-Math     & Apache-2.0 \\
    Medical-Reasoning  & Apache-2.0 \\
    UltraMedical       & MIT \\
    OSS-Instruct       & ODC-BY \\
    Code-Alpaca        & CC BY-NC 4.0 \\
    \cmidrule(lr){1-2}
    OpenHermes-2.5     & mixed \\
    Tulu-3-SFT-Mixture & ODC-BY-1.0 \\
    \bottomrule
  \end{tabular}
  \caption{Licenses of the source instruction pools.
  Top six are single-domain pools,
  bottom two multi-domain; \emph{mixed} marks OpenHermes-2.5, which
  states no single license.}
  \label{tab:licenses}
\end{table}

\section{Implementation Details}
\label{app:implementation}

This appendix gives full pseudocode for the four offline stages,
plus distributed-training engineering notes.
Following the stage definitions in the main paper, each algorithm's
\textsc{Require} line lists only the formal inputs of the corresponding
stage function. The target model and the fixed pipeline components and
configuration, including $E$, $\mathcal{T}_{\mathrm{ana}}$, $n_c$,
$n_s$, $k^*$, $k$, and $b$, are supplied by the experimental setting.

\paragraph{Agents.} Every stage reads its inputs and emits its
configuration with a single frozen instruction-tuned LLM
(DeepSeek-V4-Flash), accessed through an OpenAI-compatible API and
decoded at temperature $0.01$ (near-greedy); the same model serves
as the judge in Stage~4. The target-model metrics of Stage~3 are computed
under the target model $\theta$ being fine-tuned, not by this LLM.
Each emitted configuration is parsed and schema-checked, and a
malformed or out-of-range response is re-requested within a fixed
retry budget.

\paragraph{Tool-assisted inspection.} In Stages~2--4, the agent gathers
observations through a sandboxed Bash interface. The LLM may issue
lightweight shell queries to inspect the current data files or create
and run temporary Python scripts to compute descriptive statistics.
These operations treat the input data as read-only: temporary analysis
artifacts are kept separate, and the source records are not modified.
The LLM chooses the inspection actions according to the user instruction
and the objective of the current stage; no fixed representative subset
is constructed in advance.

\subsection{Stage 1: Domain Agent}
\label{app:stage0}

\Cref{alg:stage0} formalizes the matching step described for the
Domain Agent in the paper.

\begin{algorithm}[h]
  \caption{Domain Agent (Stage 1)}
  \label{alg:stage0}
  \begin{algorithmic}[1]
    \Require Instruction $\mathcal{I}$, raw pool $\mathcal{D}$,
      frozen LLM $\Phi$
    \State $\mathrm{mode} \gets \textsc{ModeOf}(\mathcal{I})$
      \Comment{keyword cues or LLM intent classifier}
    \If{$\mathrm{mode}$ = single-domain}
      \State \Return $\mathcal{D}$
    \EndIf
    \State $\{\mathbf{e}_i\} \gets E(\mathcal{D})$
      \Comment{embed all samples}
    \If{$|\mathcal{D}| \leq n_c$}
      \State $\{C_j\} \gets \textsc{HDBSCAN}(\{\mathbf{e}_i\})$
    \Else
      \State $\{C_j\} \gets \textsc{KMeans}(\{\mathbf{e}_i\}, k_c)$
    \EndIf
    \For{each cluster $C_j$}
      \State $R_j \gets$ random sample of $n_s$ instances from $C_j$
      \State $l_j \gets \llm(P_{lab}, R_j, \Phi)$
        \Comment{one or more category names}
    \EndFor
    \State $\mathcal{K} \gets \llm(P_{sel}, \mathcal{I},
      \{(j,l_j)\}, \Phi)$
      \Comment{matching cluster IDs}
    \State \Return $\bigcup_{j \in \mathcal{K}} C_j$
  \end{algorithmic}
\end{algorithm}

\paragraph{Clustering details.} HDBSCAN uses Euclidean distance with
\texttt{min\_cluster\_size} $=$ $50$. KMeans uses $k_c = 20$,
\texttt{n\_init} $=$ $3$, \texttt{max\_iter} $=$ $300$,
\texttt{random\_state} $=$ $42$. The clustering-switch threshold is
$n_c = 50{,}000$; above this threshold, we use KMeans to limit
clustering cost and improve scalability on larger pools. The number
of representative instances shown to the LLM per cluster is $n_s = 8$.

\subsection{Stage 2: Characteristic Agent}
\label{app:stage1}

\Cref{alg:stage1} summarizes the Characteristic Agent (Stage~2)
described in the paper.
The generated policy covers required-field checks, length bounds on
the joint text and on the instruction and output separately, an
output-to-instruction length-ratio bound, a refusal-phrase blacklist,
a noise-character ratio cap, and an English-character ratio floor.

\begin{algorithm}[h]
  \caption{Characteristic Agent (Stage 2)}
  \label{alg:stage1}
  \begin{algorithmic}[1]
    \Require Instruction $\mathcal{I}$, pool $\mathcal{D}_1$,
      frozen LLM $\Phi$
    \State $\mathcal{O}_2 \gets
      \operatorname{Inspect}(\mathcal{I},\mathcal{D}_1;
      \Phi,\mathcal{T}_{\mathrm{ana}})$
    \State $\mathcal{R} \gets
      \llm(P_{char},\mathcal{I},\mathcal{O}_2,\Phi)$
      \Comment{generate the filtering policy}
    \State $\mathcal{D}_2 \gets \emptyset$
    \For{each $(x, y, t)$ in $\mathcal{D}_1$ with full text $t$}
      \If{required fields \texttt{instruction}/\texttt{input}/\texttt{output}/\texttt{text}
            all present \textbf{and}
          $L_{\min} \leq |t| \leq L_{\max}$ \textbf{and}
          $|x| \geq L_x$ \textbf{and} $|y| \geq L_y$ \textbf{and}
          $|y| \geq \rho_{\min} |x|$ \textbf{and}
          $y \notin \textsc{RefusalPatterns}$ \textbf{and}
          $\textsc{NoiseRatio}(t) \leq \tau_n$ \textbf{and}
          $\textsc{EnRatio}(t) \geq \tau_e$}
        \State $\mathcal{D}_2 \gets \mathcal{D}_2 \cup \{(x, y, t)\}$
      \EndIf
    \EndFor
    \State \Return $\mathcal{D}_2$
  \end{algorithmic}
\end{algorithm}

\paragraph{Threshold defaults.} Joint-text length bounds $L_{\min} =
40$, $L_{\max} = 4096$; per-field minimums $L_x = 5$ (instruction),
$L_y = 15$ (output); output/instruction ratio $\rho_{\min} = 0.2$;
noise-character ratio cap $\tau_n = 0.15$; English-character ratio
floor $\tau_e = 0.30$. \textsc{RefusalPatterns} covers common
alignment-refusal phrases (e.g., \texttt{"I'm sorry"}, \texttt{"I
cannot"}, \texttt{"As an AI language model"}). The agent may override
any default by emitting a \texttt{policy} dict at the infer step.

\subsection{Stage 3: Informativeness Agent}
\label{app:stage2}

\Cref{alg:stage2} summarizes the Informativeness Agent (Stage~3)
described in the paper.

\begin{algorithm}[h]
  \caption{Informativeness Agent (Stage 3)}
  \label{alg:stage2}
  \begin{algorithmic}[1]
    \Require Instruction $\mathcal{I}$, pool $\mathcal{D}_2$,
      frozen LLM $\Phi$
    \State $\mathcal{O}_3 \gets
      \operatorname{Inspect}(\mathcal{I},\mathcal{D}_2;
      \Phi,\mathcal{T}_{\mathrm{ana}})$
    \State $\mathbf{w}=[w^1,\dots,w^{m_1}] \gets
      \llm(P_{inf},\mathcal{I},\mathcal{O}_3,\Phi)$
      \Comment{$\sum_{j=1}^{m_1} w^j \!=\! 1$, $w^j \!\geq\! 0$}
    \For{$j=1,\dots,m_1$}
      \State Compute $c_i^j$ for all $d_i \in \mathcal{D}_2$;
      normalize and adjust polarity
    \EndFor
    \State $S(\mathcal{I},\mathcal{D}_2,\Phi) \gets
      \{\sum_{j=1}^{m_1} w^j c_i^j \mid d_i \in \mathcal{D}_2\}$
    \State \Return the top $k^*$ samples in $\mathcal{D}_2$ ranked
    by $S$
  \end{algorithmic}
\end{algorithm}

\subsection{Stage 4: Quality Agent}
\label{app:stage3}

\Cref{alg:stage3} summarizes the Quality Agent (Stage~4) described in
the paper.

\begin{algorithm}[h]
  \caption{Quality Agent (Stage 4)}
  \label{alg:stage3}
  \begin{algorithmic}[1]
    \Require Instruction $\mathcal{I}$, pool $\mathcal{D}_3$,
      frozen LLM $\Phi$
    \State $\mathcal{O}_4 \gets
      \operatorname{Inspect}(\mathcal{I},\mathcal{D}_3;
      \Phi,\mathcal{T}_{\mathrm{ana}})$
    \State $\mathcal{R}_{qual}=\{(r^j,z^j)\}_{j=1}^{m_2} \gets
      \llm(P_{rub},\mathcal{I},\mathcal{O}_4,\Phi)$
      \Comment{$\sum_{j=1}^{m_2} z^j = 1$}
    \For{each batch $\mathcal{B} \subset \mathcal{D}_3$ of size $b$ in parallel}
      \State $\{[q_i^1,\dots,q_i^{m_2}] \mid d_i \in \mathcal{B}\}
        \gets \llm(P_{cri}(\mathcal{R}_{qual}),\mathcal{B},\Phi)$
      \Comment{integer in $[1,5]$}
    \EndFor
    \State $Q(\mathcal{I},\mathcal{D}_3,\Phi) \gets
      \{\sum_{j=1}^{m_2} z^j q_i^j \mid d_i \in \mathcal{D}_3\}$
    \State \Return the top $k$ samples in $\mathcal{D}_3$ ranked by $Q$
  \end{algorithmic}
\end{algorithm}

\paragraph{Example rubric.} \Cref{tab:stage3-rubric} presents a
representative rubric for a math-reasoning task instruction
$\mathcal{I}$, consistent with the rubric-generation procedure in
Stage~4.
The four dimensions, their weights, and their descriptions are all
generated from $\mathcal{I}$ and the tool-assisted observations
$\mathcal{O}_4$; none is drawn from a
fixed list, and the weights sum to one.

\begin{table*}[!t]
  \centering
  \small
  \setlength{\tabcolsep}{5pt}
  \begin{tabular}{@{}p{3.2cm} c p{\dimexpr\textwidth-5.0cm\relax}@{}}
    \toprule
    Dimension & $z^j$ & Representative description \\
    \midrule
    \texttt{mathematical\_\allowbreak relevance} & $0.35$ &
    Does the sample involve genuine mathematical reasoning? Score high
    for: multi-step problem solving, symbolic manipulation, proofs,
    formula reasoning, word problems with quantitative reasoning, and
    computation requiring chain-of-thought. Score low for: code-only
    without math, general science without explicit math derivation,
    medical/clinical content, humanities, chitchat, or outputs that
    fail to address the math in the instruction. \\
    \texttt{logical\_\allowbreak rigor} & $0.30$ &
    Is the mathematical reasoning logically consistent, step-by-step,
    and free of contradictions? Does the output show rigorous
    derivation rather than hand-waving? Score high for clear
    step-by-step derivations, proper use of mathematical notation, and
    logically sound conclusions. \\
    \texttt{instruction\_\allowbreak following} & $0.20$ &
    Does the output correctly and completely address the mathematical
    instruction? Score high when the answer is accurate, relevant, and
    directly responds to the math problem posed. Score low for
    off-topic responses, incorrect answers, or outputs that ignore the
    mathematical nature of the question. \\
    \texttt{educational\_\allowbreak value} & $0.15$ &
    Does the sample have high educational value for math training?
    Score high for: clear exposition of mathematical concepts,
    well-structured derivations that serve as good demonstrations,
    non-trivial problems that teach reasoning patterns. Score low for
    trivial arithmetic, one-step lookup answers, or overly simplistic
    problems. \\
    \bottomrule
  \end{tabular}
  \caption{Example Stage~4 rubric emitted at run time. The
  judging dimensions, weights, and descriptions that \OurMethod's Stage~4
  agent generated from a math-reasoning task instruction $\mathcal{I}$ in the
  Quality Agent stage. The dimensions are produced at run time
  rather than chosen from a fixed list; each candidate is then rated
  $1$--$5$ on every dimension and ranked by the weighted sum.}
  \label{tab:stage3-rubric}
\end{table*}

\paragraph{Batched scoring.} Each round consumes
\texttt{samples\_per\_round} $=$ $50$ candidates split into
\texttt{num\_batches} $=$ $5$ batches of size $b = 10$, with the
five batches dispatched concurrently via a thread pool. Each
$P_{cri}$ (see \Cref{app:prompts}) asks the LLM to return
a JSON array of objects $\{$\texttt{sample\_id}, \texttt{scores}:
\{d: \texttt{int}\}, \texttt{reasoning}, \texttt{flagged}$\}$,
calibrated comparatively across the batch. JSON parse failures are
caught and produce empty results for that batch (no retry); the
remaining round continues uninterrupted.

\subsection{Distributed-Training Engineering}
\label{app:engineering}

\paragraph{Training framework.} Fine-tuning uses full-parameter SFT
under DeepSpeed ZeRO-3~\citep{rajbhandari2020zero} with bfloat16
mixed precision. The frozen LLM used to configure all four stages, and
additionally used for judging in Stage~4, is accessed through an
OpenAI-compatible API.

\paragraph{Evaluator chat-template handling.} Some evaluation
harnesses inadvertently emit chat-template tokens (e.g.,
\texttt{<|im\_start|>}) inside the prompt fed to target models, which
inflates apparent baseline scores by leaking instruction-tuning
artefacts. We override the tokenizer's \texttt{chat\_template} field
with a minimal raw-prompt template across all three target models,
ensuring consistent zero-shot evaluation.

\section{Baseline Implementation Notes}
\label{app:baseline-impl}


\paragraph{Full Pool.} In the single-domain experiments, Full Pool
fine-tunes each target model on the entire uncurated source pool under
the same training and evaluation protocol as the selection methods. It
is not included in the multi-domain comparison for the reason described
in the main paper.

\paragraph{Random.} Random uniformly samples 10K examples without
replacement to match \OurMethod's subset budget. We reuse the
\texttt{random\_state} $=$ $42$ seed adopted by KMeans in Stage~1 for
reproducibility.

\paragraph{SuperFiltering.} \citet{li2024superfiltering} compute the
IFD ratio of conditional to unconditional cross-entropy with a
GPT-2-style proxy model and retain the highest-scoring samples until
the budget is met.

\paragraph{Select2Reason.} Because the authors' implementation is not
publicly available, we independently reimplement Select2Reason following
the procedure described by \citet{yang2025select2reason}. For each
question, we estimate its difficulty from the empirical solve rate
obtained through Monte-Carlo rollouts of the corresponding target model
and use these rollout-derived scores to train a reward model for scalable
difficulty estimation. We normalize each reasoning trace by removing
exactly repeated reasoning steps and rank the resulting traces by their
effective lengths. Following the original method, the difficulty and
length rankings are combined as
$\mathrm{rank}_j = w \cdot \mathrm{rank}_d +
(1-w) \cdot \mathrm{rank}_l$, with $w=0.25$, and the highest-ranked
10K samples are retained to match the selection budget used by all
methods.

\paragraph{MIG.} \citet{chen2025mig} fix several pieces of their
selection objective by paper convention: tag-edge weights, the concave
information function, and the propagation factor. The semantic tags
themselves are produced in practice by the open-source
\textsc{InsTagger} model (rather than re-annotating with the
\textsc{InsTag} taxonomy from scratch).

\paragraph{SelectIT.} \citet{liu2024selectit} aggregate token-level
probability disparities under $K$ rating prompts into a token-grain
score, then into a sentence-grain
$\mathrm{Avg}/(1 \!+\! \alpha \mathrm{Std})$, then into a model-grain
combination across multiple foundation LLMs weighted by parameter
count.

\paragraph{DEFT.} \citet{DEFT} rank each sample by the score
$\mathit{complexity}\times\mathit{quality} +
\mathit{quality}\times\mathit{knowledge}$ (each term min-max normalized),
where complexity and quality come from distilled DeBERTa-v3 rankers and
knowledge is a target-model min-$k\%$ negative log-likelihood (the one
model-aware signal). They then walk down this ranking and greedily keep a
sample only if its cosine similarity to the already-selected set (on
\texttt{e5-large-v2} embeddings) stays below a threshold, until the 10K
budget is filled; diversity is thus a selection constraint, not a term in
the score. DEFT is a task-agnostic quality selector, so we run it on the
single-domain pools only.

\paragraph{SuperFiltering (Tag) and SuperFiltering (Sim).} Both are
two-stage multi-domain baselines. SuperFiltering (Tag) uses source tags
to identify target-domain candidates, whereas SuperFiltering (Sim)
ranks candidates by their BGE-M3 embedding similarity to target
examples. SuperFiltering then selects the final 10K subset from the
resulting candidate pool.

\paragraph{DSIR and TSDS.} Both are target-guided static baselines
evaluated in the paper's multi-domain setting.
They use a reference set of target examples without a separate
downstream quality scorer.
\citet{xie2023dsir} hash each example's unigrams and bigrams into
10K buckets, fit a bag-of-$n$-grams model on the target examples and
another on the raw pool, score each candidate by the resulting
target-to-raw likelihood ratio, and importance-resample the budget
without replacement using Gumbel top-$k$. \citet{liu2024tsds} optimize
probability transport from target query examples to pool candidates,
combining an optimal-transport distribution-alignment cost with a
diversity regularizer. In their KNN-KDE instantiation, probability mass
is assigned to each query's candidate neighborhood in inverse proportion
to estimated local density, reducing near-duplicate oversampling; samples
are then drawn with replacement from the resulting categorical
distribution.

\paragraph{Claude Code.} Claude Code uses Claude Opus 4.7 as its
backend model and receives the same natural-language selection
instruction and 10K subset budget as \OurMethod. It operates as a
cost-aware single-pass curator and returns the selected subset without
the explicit four-stage cascade used by \OurMethod.

\section{Reproducibility Details}
\label{app:reproducibility}

\paragraph{Models.} We evaluate two base models, Qwen2.5-7B and
Llama-3.1-8B, together with one instruction-tuned model,
Olmo-3-7B-Instruct; all three models are in the 7--8B-parameter range.
The frozen LLM agent and its decoding configuration are described in
Appendix~\ref{app:implementation}.

\paragraph{Fine-tuning protocol.} Each fine-tuning run uses
$3$ epochs and full-parameter SFT under DeepSpeed
ZeRO-3~\citep{rajbhandari2020zero} with bfloat16 mixed precision.
Before the large-scale evaluation, a broader pilot sweep on several
datasets and target models was used to retain three learning rates that
performed well across most pilot settings:
$\{1\!\times\!10^{-6}, 5\!\times\!10^{-6}, 1\!\times\!10^{-5}\}$.
In each formal experimental setting, every compared method is run at
all three rates. For each rate, we average each method's setting-level
score (the arithmetic mean over the benchmarks in its domain) across
all methods and select the rate with the highest average; all methods
in that setting are reported at this same shared rate. No method-specific
learning-rate tuning is performed.
The selection methods use a 10K subset, while
\emph{Full Pool} uses the entire uncurated source pool. Reported
scores are from the final (third) epoch.

\paragraph{Runs per setting.} Each setting is reported on a single
seed of $42$, applied to Python sampling during selection,
training-data ordering, NumPy, PyTorch, CUDA, and KMeans. The training
entry point records the seed in each run's configuration. Hosted-LLM
services and some distributed CUDA kernels may nevertheless not be
bitwise deterministic, and multi-seed variance is not quantified.


\paragraph{Hardware.}
The agentic curation pipeline (Stage~1--4, including the Informativeness
Agent and the bge-m3 clustering) runs on a single node
with $8\times$ NVIDIA RTX 5880 Ada Generation GPUs ($48$~GB each;
$2\times$ Intel Xeon Gold 6530, $503$~GB RAM; Ubuntu 22.04). The
Informativeness Agent (Stage~3) runs the target model on
these GPUs, while \OurMethod's agent reasoning calls use the hosted
DeepSeek-V4-Flash API. Full-parameter SFT ($3$ epochs, DeepSpeed
ZeRO-3, and bfloat16, following the paper's experimental
setup) is run on a separate node with
$8\times$ NVIDIA H200 GPUs ($143{,}771$~MiB each; driver 570.133.20),
an Intel Xeon Processor,
$64$~GiB RAM visible to the job, and Ubuntu 24.04.1 LTS.

\paragraph{Software.}
\Cref{tab:software} lists the main library versions used in the
experiments.

\begin{table}[!t]
  \centering
  \small
  \setlength{\tabcolsep}{4pt}
  \begin{tabular}{@{}ll@{}}
    \toprule
    Library & Version \\
    \midrule
    Python                & 3.12 \\
    PyTorch               & 2.10.0 (CUDA 12.8) \\
    Transformers          & 4.57.6 \\
    Tokenizers            & 0.22.2 \\
    vLLM                  & 0.19.0 \\
    DeepSpeed             & 0.18.9 \\
    Accelerate            & 1.13.0 \\
    Datasets              & 4.8.4 \\
    sentence-transformers & 5.4.1 \\
    FAISS                 & 1.13.2 \\
    NumPy                 & 2.2.6 \\
    LangGraph             & 1.1.6 \\
    langchain-openai      & 1.1.12 \\
    openai                & 2.31.0 \\
    \bottomrule
  \end{tabular}
  \caption{Experimental software versions.}
  \label{tab:software}
\end{table}

\section{LLM Prompts}
\label{app:prompts}

This section reproduces the six prompt templates used by the method,
one table per template:
$P_{lab}$ (\Cref{tab:prompt-lab}) and $P_{sel}$
(\Cref{tab:prompt-sel}) for the Domain Agent, $P_{char}$
(\Cref{tab:prompt-char}) for the Characteristic Agent, $P_{inf}$
(\Cref{tab:prompt-inf}) for the Informativeness Agent, and
$P_{rub}$ (\Cref{tab:prompt-rub}) together with $P_{cri}$
(\Cref{tab:prompt-cri}) for the Quality Agent. Curly braces mark
slots filled at run time. Note that $P_{cri}$ is not a fixed
template: its evaluation-dimension block is instantiated at run
time with the rubric emitted under $P_{rub}$, so only its scaffold
is fixed and reproduced here (an example run-time rubric is given
in \Cref{tab:stage3-rubric}). All prompts are sent to the same
OpenAI-compatible API endpoint, and every template that requests a
configuration instructs the LLM to return strictly-valid JSON.
For Stages~2--4, tool-assisted inspection is completed before the
corresponding decision prompt is invoked. The resulting observations
$\mathcal{O}_2$, $\mathcal{O}_3$, and $\mathcal{O}_4$ fill the
run-time slots shown below.

\begin{table*}[!t]
  \centering
  \small
  \begin{tabular}{@{}p{\dimexpr\textwidth-0.4cm\relax}@{}}
    \toprule
    \textbf{Prompt template $P_{lab}$ --- Domain Agent: cluster labeling} \\
    \midrule
    You are the Domain Agent, the first stage of \OurMethod's
    data-curation cascade. The data pool has been partitioned into
    clusters by embedding-based clustering, and you are shown several
    samples drawn at random from one cluster. Your job is to produce a
    short list of category names describing the primary semantic
    content of this cluster. \\[4pt]
    \textbf{Rules:} \\
    \textbullet\ Return a JSON list containing one or more short
    category names in English (e.g.\ ``cardiology'', ``organic
    chemistry'', ``linear algebra'', ``Python programming'', or
    ``classical mechanics''). \\
    \textbullet\ Be specific but not overly narrow. Use a single broad
    category when the samples span closely related areas, and use
    multiple categories only when distinct subjects are jointly
    represented. \\
    \textbullet\ If the samples are too diverse to fit one discipline,
    return \texttt{["mixed"]}. \\[4pt]
    Respond with ONLY the JSON list, nothing else. \\[4pt]
    \emph{[Input]} \{samples drawn at random from one cluster:
    instruction and truncated output, up to 8 samples\} \\
    \bottomrule
  \end{tabular}
  \caption{Cluster-labeling prompt $P_{lab}$. Applied once
  per cluster; the category-name lists it produces form the label space
  read by $P_{sel}$ (\Cref{tab:prompt-sel}). The prompt sees only
  samples, never the user instruction $\mathcal{I}$, so labeling is
  independent of the selection target.}
  \label{tab:prompt-lab}
\end{table*}

\begin{table*}[!t]
  \centering
  \small
  \begin{tabular}{@{}p{\dimexpr\textwidth-0.4cm\relax}@{}}
    \toprule
    \textbf{Prompt template $P_{sel}$ --- Domain Agent: cluster selection} \\
    \midrule
    You are the Domain Agent, the first stage of \OurMethod's
    data-curation cascade. The data pool has been partitioned into
    clusters, and each cluster is paired with its category-name list.
    Given the user's data-selection instruction, your job is to decide
    which clusters match the user's target domain; only the matching
    clusters are passed to the next stage. \\[4pt]
    \textbf{Context:} \\
    \textbullet\ User Instruction: \{user instruction $\mathcal{I}$\} \\
    \textbullet\ Cluster Labels:
    \{$(1,l_1), \dots, (m,l_m)$\} \\[4pt]
    \textbf{Rules:} \\
    \textbullet\ Identify the target domain(s) from the user
    instruction. Treat them as BROAD disciplines (e.g.\
    ``mathematics'', ``medicine''), not narrow sub-fields. \\
    \textbullet\ Keep a cluster if any category in its label list
    belongs to (or is a
    sub-field of) a target domain. For example, if the target is
    ``mathematics'': ``linear algebra'' $\rightarrow$ keep;
    ``calculus'' $\rightarrow$ keep; ``statistics'' $\rightarrow$ keep
    (mathematical discipline); ``python programming'' $\rightarrow$
    discard; ``organic chemistry'' $\rightarrow$ discard. \\
    \textbullet\ If the instruction names no target domain (e.g.\
    ``select general high-quality samples''), keep ALL clusters. \\[4pt]
    Output a JSON object containing only the IDs of matching clusters:
    \\
    \texttt{\{"matching\_cluster\_ids": [1, 4, 7]\}} \\
    Return ONLY the JSON object. \\
    \bottomrule
  \end{tabular}
  \caption{Cluster-selection prompt $P_{sel}$. Reads the
  user instruction $\mathcal{I}$ together with the category-name lists
  produced under $P_{lab}$ and returns the matching cluster-ID
  set $\mathcal{K}$; the matched clusters form $\mathcal{D}_1$.}
  \label{tab:prompt-sel}
\end{table*}

\begin{table*}[!t]
  \centering
  \small
  \begin{tabular}{@{}p{\dimexpr\textwidth-0.4cm\relax}@{}}
    \toprule
    \textbf{Prompt template $P_{char}$ --- Characteristic Agent: rule-policy generation} \\
    \midrule
    You are the Characteristic Agent, the second stage of \OurMethod's
    data-curation cascade. Given the user's data-selection instruction
    and tool-assisted observations of the current data pool, your job
    is to configure a rule-based filtering policy tailored to the user's target; a
    deterministic filter then applies your policy to every sample,
    with no further model calls. \\[4pt]
    \textbf{Context:} \\
    \textbullet\ User Instruction: \{user instruction $\mathcal{I}$\} \\
    \textbullet\ Tool-assisted Observations:
    \{observations $\mathcal{O}_2$ collected from $\mathcal{D}_1$\}
    \\[4pt]
    \textbf{Output:} Emit a policy dict containing ALL keys below ---
    do not add, remove, or rename any key: \\
    \textbullet\ \texttt{min\_total\_len} (int): Minimum total sample
    length. \\
    \textbullet\ \texttt{max\_total\_len} (int): Maximum total sample
    length. \\
    \textbullet\ \texttt{min\_inst\_len} (int): Minimum instruction
    length. \\
    \textbullet\ \texttt{min\_out\_len} (int): Minimum output
    length. \\
    \textbullet\ \texttt{out\_inst\_ratio\_min} (float): Minimum
    output/instruction length ratio. \\
    \textbullet\ \texttt{refusal\_phrases} (list[str]): Phrases
    indicating refusals, e.g.\ \texttt{["sorry", "I cannot answer",
    "As an AI"]}. \\
    \textbullet\ \texttt{noise\_regex} (str): Regex matching noise
    characters, e.g.\
    \texttt{[\^{}\textbackslash u4e00-\textbackslash
    u9fa5\textbackslash w\textbackslash s,.;!?()/\textbackslash
    uff0c\textbackslash u3002\textbackslash uff01\textbackslash
    uff1f]}. \\
    \hspace*{1em}IMPORTANT: Adjust for domain-specific data: \\
    \hspace*{1em}$\ast$ Math: allow $\sum$, $\int$, $\surd$, $\infty$,
    $\pi$, +, -, *, \^{}, \_, etc. \\
    \hspace*{1em}$\ast$ Code: allow \{\}, [], $<$$>$, $|$, \&, \#,
    etc. \\
    \textbullet\ \texttt{max\_noise\_ratio} (float): Maximum
    noise-to-total ratio. \\
    \textbullet\ \texttt{require\_en} (bool): Whether to require
    English content. \\
    \textbullet\ \texttt{min\_en\_ratio} (float): Minimum English
    character ratio. \\[4pt]
    Return the policy as a JSON object. \\
    \bottomrule
  \end{tabular}
  \caption{Rule-policy prompt $P_{char}$. Emits the
  characteristic-level rule set $\mathcal{R}$; a deterministic filter
  then applies the policy to every sample of $\mathcal{D}_1$ without
  further LLM calls.}
  \label{tab:prompt-char}
\end{table*}

\begin{table*}[!t]
  \centering
  \small
  \begin{tabular}{@{}p{\dimexpr\textwidth-0.4cm\relax}@{}}
    \toprule
    \textbf{Prompt template $P_{inf}$ --- Informativeness Agent: metric-weight allocation} \\
    \midrule
    You are the Informativeness Agent, the third stage of \OurMethod's
    data-curation cascade. Given the user's data-selection instruction
    and tool-assisted observations of the current data pool, your job
    is to allocate weights over four target-model metrics according to
    the user's target; every sample is then scored by the weighted sum
    of its normalized metrics, and the top-ranked fraction is kept. \\[4pt]
    \textbf{Context:} \\
    \textbullet\ User Instruction: \{user instruction $\mathcal{I}$\} \\
    \textbullet\ Tool-assisted Observations:
    \{observations $\mathcal{O}_3$ collected from $\mathcal{D}_2$\}
    \\[4pt]
    \textbf{Metrics} (each computed with the target model and
    min-max normalized to [0, 1]; after normalization, higher = more
    desirable). For each metric, the selection intent it serves and
    the instruction cues that should raise its weight: \\
    \textbullet\ \texttt{loss}: NLL of the reference response ---
    measures the information gap between the model and the data.
    Serves selecting samples with large knowledge gaps and high
    information density; raise its weight when the instruction
    emphasizes hard, rare, or specialized samples. \\
    \textbullet\ \texttt{entropy}: Token-level entropy --- measures
    model uncertainty / knowledge blind spots. Serves covering the
    model's uncertain or blind-spot areas; raise its weight when the
    instruction emphasizes the model's weak areas. \\
    \textbullet\ \texttt{drift}: Semantic drift between the model's
    own output and the reference --- measures the
    instruction-following alignment gap. Serves selecting samples
    whose output form or reasoning style departs most from the
    model's current behavior; raise its weight when the instruction
    emphasizes a desired output form such as step-by-step or
    chain-of-thought reasoning. \\
    \textbullet\ \texttt{mean\_diff}: Parameter-change magnitude after
    one gradient step (ResoFilter) --- measures data stability.
    Serves selecting clean, low-noise samples the model can learn
    from stably; raise its weight when the instruction emphasizes
    reliable, stable, or clean data. \\[4pt]
    \textbf{Output:} Emit a weight dict over the four metrics by
    matching the requirements expressed in the instruction to the
    metrics above. Weights MUST sum to 1.0. Avoid assigning a weight
    of exactly 0 to any metric: if a metric is of low relevance to
    the user's target, assign it a small but non-zero weight (e.g.\
    0.05) rather than 0. \\[4pt]
    Return ONLY the JSON dict, e.g.: \\
    \texttt{\{"loss": 0.2, "entropy": 0.15, "drift": 0.25,
    "mean\_diff": 0.4\}} \\
    \bottomrule
  \end{tabular}
  \caption{Metric-weight prompt $P_{inf}$. Emits the weight
  vector $[w^1, \dots, w^{m_1}]$ over the four target-model metrics; the
  metrics themselves are computed under the target model and combined
  by the weighted sum that ranks $\mathcal{D}_2$.}
  \label{tab:prompt-inf}
\end{table*}

\begin{table*}[!t]
  \centering
  \small
  \begin{tabular}{@{}p{\dimexpr\textwidth-0.4cm\relax}@{}}
    \toprule
    \textbf{Prompt template $P_{rub}$ --- Quality Agent: rubric generation} \\
    \midrule
    You are the Quality Agent, the fourth and final stage of
    \OurMethod's data-curation cascade. Given the user's
    data-selection instruction and observations gathered through
    tool-assisted inspection of the current data pool, your job is to
    design a judging rubric --- a set of quality dimensions with
    weights --- tailored to the user's target. An LLM
    judge then scores every sample against your rubric, and the
    top-ranked samples form the final curated subset. \\[4pt]
    \textbf{Context:} \\
    \textbullet\ User Instruction: \{user instruction $\mathcal{I}$\} \\
    \textbullet\ Tool-assisted Observations:
    \{observations $\mathcal{O}_4$ collected from $\mathcal{D}_3$\}
    \\[4pt]
    \textbf{Output:} Emit a dimensions dict mapping each dimension
    name to a description and a weight. Weights MUST sum to 1.0.
    Choose dimensions and weights appropriate to the dataset and the
    user's goal --- when the instruction implies specific
    requirements, prefer targeted dimensions over generic ones. \\[4pt]
    Example: \\
    \texttt{\{"instruction\_following": \{"description": "Does the
    output strictly follow the given instruction and constraints?",
    "weight": 0.4\},} \\
    \texttt{~~"logical\_rigor": \{"description": "Is the output
    logically consistent and free of contradictions?", "weight":
    0.4\},} \\
    \texttt{~~"information\_density": \{"description": "Is the output
    comprehensive and covers all aspects of the request?", "weight":
    0.2\}\}} \\[4pt]
    Return ONLY the JSON dict. \\
    \bottomrule
  \end{tabular}
  \caption{Rubric-generation prompt $P_{rub}$. Emits the
  open-vocabulary judging rubric (dimension names, descriptions, and
  weights) from the user instruction $\mathcal{I}$ and tool-assisted
  observations $\mathcal{O}_4$; the rubric instantiates the judging prompt $P_{cri}$
  (\Cref{tab:prompt-cri}). A representative rubric instance is shown in
  \Cref{tab:stage3-rubric}.}
  \label{tab:prompt-rub}
\end{table*}

\begin{table*}[!t]
  \centering
  \small
  \begin{tabular}{@{}p{\dimexpr\textwidth-0.4cm\relax}@{}}
    \toprule
    \textbf{Judging prompt $P_{cri}$ --- Quality Agent: batch scoring
    (scaffold; instantiated at run time with the rubric from
    $P_{rub}$)} \\
    \midrule
    \textbf{Role}: Professional Instruction-Data Quality Auditor. \\
    \textbf{Task}: Evaluate the quality of a batch of dataset samples
    \textbf{comparatively}, based on the quality dimensions below. \\[4pt]
    You will receive \{batch\_size\} samples at once. By seeing them
    side by side, you can calibrate your ratings --- use the full 1-5
    scale, differentiating strong samples from weak ones within this
    batch. \\[4pt]
    \textbf{Evaluation Dimensions:} \\
    \{rubric emitted under $P_{rub}$: one line per dimension --- name,
    description, weight\} \\[4pt]
    \textbf{Rating Scale (1-5):} \\
    1: Terrible (Completely failed) \\
    2: Poor (Major issues) \\
    3: Fair (Acceptable but needs improvement) \\
    4: Good (Minor issues) \\
    5: Excellent (Perfect) \\[4pt]
    \textbf{Output Format:} \\
    Return a JSON \textbf{array} with exactly \{batch\_size\} objects,
    one per sample, in the same order as the input: \\
    \texttt{[\{"sample\_id": <id\_from\_input>, "scores":
    \{"<dimension\_name>": <1-5>, ...\}, "reasoning": "Brief
    explanation of the scores (1-2 sentences)", "flagged":
    <true/false>\}, ...]} \\[4pt]
    \textbf{Important:} \\
    \textbullet\ Compare samples against each other to maintain
    consistent, objective scoring across the batch. \\
    \textbullet\ Use the full rating range: if there are clearly
    better and worse samples, reflect that in the scores. \\
    \textbullet\ If a sample contains harmful, illegal, or NSFW
    content, set ``flagged'' to true and give low scores. \\
    \textbullet\ Return ONLY the JSON array with no additional
    text. \\[4pt]
    \emph{[Input]} \{batch of samples, each rendered as ``---~Sample
    $\langle$id$\rangle$~--- Instruction: \dots\ Input: \dots\
    Output: \dots''\} \\
    \bottomrule
  \end{tabular}
  \caption{Batch-scoring prompt $P_{cri}$. The fixed
  scaffold of the LLM-judge prompt; its evaluation-dimension block is
  filled at run time with the rubric emitted under $P_{rub}$, so
  $P_{cri}$ is a run-time product rather than a fixed template.
  Candidates are scored in batches and ranked by the weighted sum of
  their per-dimension ratings.}
  \label{tab:prompt-cri}
\end{table*}

\section{Broader Evaluation and Extensions}
\label{app:extensions}

\paragraph{Scaling across model families.}
Our evaluation covers three target models at the 7B--8B scale. Future
work can extend this evaluation to larger models and a broader range of
model architectures.

\paragraph{Broader dynamic-curator comparisons.}
The Claude Code baseline represents a practical cost-aware single-pass
curation configuration. Future evaluations can broaden this comparison
across curator models, prompt designs, and inference budgets.

\paragraph{Adaptive orchestration.}
Building on instruction-conditioned configuration within each stage,
future work can extend adaptation to stage composition and routing,
allowing the cascade structure itself to respond to the instruction and
observed data pool.